\documentclass{article} % For LaTeX2e
\usepackage[preprint]{colm2026_conference}

\usepackage{microtype}
\usepackage{hyperref}
\usepackage{url}
\usepackage{booktabs}
\usepackage{algorithm}
\usepackage{algpseudocode}
\usepackage{graphicx}
\usepackage{multirow}
\usepackage[most]{tcolorbox}
\usepackage{float}
\usepackage{titlesec}

\usepackage{amssymb}
\usepackage{xcolor}
\newcommand{\cmark}{{\color{green!60!black}\checkmark}}
\newcommand{\xmark}{{\color{red!60!black}$\times$}}

\usepackage{amsmath,amsfonts,bm}

\usepackage{cleveref}
\usepackage{wrapfig}
\usepackage{caption}

\usepackage[most]{tcolorbox}
\usepackage[table]{xcolor}
\usepackage{listings}
\usepackage[T1]{fontenc}
\usepackage{subcaption}
\usepackage{makecell}

\def\eqref#1{equation~\ref{#1}}
\def\1{\bm{1}}

\DeclareMathAlphabet{\mathsfit}{\encodingdefault}{\sfdefault}{m}{sl}
\SetMathAlphabet{\mathsfit}{bold}{\encodingdefault}{\sfdefault}{bx}{n}

\newcommand{\KL}{D_{\mathrm{KL}}}

\def\yh#1{}
\def\jiarui#1{}
\def\ap#1{}
\def\liam#1{}
\def\edward#1{}

\definecolor{pink}{HTML}{FD92A5}
\definecolor{darkyellow}{HTML}{E9B34F}
\definecolor{lightyellow}{HTML}{FDECA9}
\definecolor{bluegreen}{HTML}{039594}
\definecolor{darkbluegreen}{HTML}{00879B}
\definecolor{vividblue}{HTML}{66C5CC}
\definecolor{lightblue}{HTML}{DFF3F6}

\newcommand{\ours}{\textsc{SD-Zero}}
\newcommand{\sft}{\textsc{SRT}}
\newcommand{\method}{Self-Distillation Zero}

\newcommand{\stopgrad}{\text{StopGrad}}

\newcommand{\lsrt}{$\mathcal{L}_{\text{\sft{}}}$}
\newcommand{\lgen}{$\mathcal{L}_{\text{generation}}$}
\newcommand{\lrev}{$\mathcal{L}_{\text{revision}}$}

\newcommand{\qwen}{Qwen3-4B-Instruct}
\newcommand{\olmo}{Olmo-3-7B-Instruct}
\newcommand{\opsd}{Self-Distillation}
\newcommand{\openmath}{OpenR1-Math}
\newcommand{\cf}{Codeforces}

\usepackage{lineno}
\usepackage{fontawesome5}
\usepackage{hyperref}

\definecolor{darkblue}{rgb}{0, 0, 0.5}
\hypersetup{colorlinks=true, citecolor=darkblue, linkcolor=darkblue, urlcolor=darkblue}

\title{Self-Distillation Zero: Self-Revision Turns Binary Rewards into Dense Supervision}
\author{Yinghui He$^1$ \quad Simran Kaur$^1$ \quad Adithya Bhaskar$^1$ \quad Yongjin Yang$^2$ \quad
 \textbf{Jiarui Liu}$^3$ \\ \textbf{Narutatsu Ri}$^1$\quad \textbf{Liam Fowl}$^1$ \quad \textbf{Abhishek Panigrahi}$^1$ \quad \textbf{Danqi Chen}$^1$ \quad \textbf{Sanjeev Arora}$^1$   \\
$^1$Princeton University \quad
$^2$University of Toronto \quad
$^3$Carnegie Mellon University\\
\texttt{yh0068@princeton.edu} \\
}

\newcommand{\yj}[1]{}
\newcommand{\sk}[1]{}
\newcommand{\danqi}[1]{}
\renewcommand{\sk}[1]{}

\begin{document}

% \titlespacing*{\subsection}
% {0pt}{0.4em}{0.4em}

\ifcolmsubmission
\linenumbers    
\fi

\maketitle

\begin{abstract}
\looseness-1Current post-training methods in verifiable settings fall into two categories. Reinforcement learning (RLVR) relies on binary rewards, which are broadly applicable and powerful, but provide only sparse supervision during training. Distillation provides dense token-level supervision, typically obtained from an external teacher or using high-quality demonstrations. Collecting such supervision can be costly or unavailable.
We propose \textbf{\method{}} (\ours{}{}), a method that is substantially more training sample-efficient
than RL and doesn't require an external teacher or high-quality demonstrations. \ours{}{} trains a \emph{single model} to play two roles: a \emph{Generator}, which produces an initial response, and a \emph{Reviser}, which conditions on that response and its binary reward to produce an improved response. We then perform on-policy self-distillation to distill the reviser into the generator, using the reviser’s token distributions conditioned on the generator’s response and its reward as supervision. In effect, \ours{}{} trains the model to \textit{transform binary rewards into dense token-level self-supervision.}

\looseness-1On math and code reasoning benchmarks with \qwen{} and \olmo{}, \ours{}{} improves performance by at least $10\%$ over the base models and outperforms strong baselines, including Rejection Fine-Tuning (RFT), GRPO, and Self-Distillation Fine-Tuning (SDFT), under the same question set and training sample budget.
Extensive ablation studies show two novel characteristics of our proposed algorithm: (a) \emph{token-level self-localization}, where the reviser can identify the key tokens that need to be revised in the generator's response based on reward,  and (b) \emph{iterative self-evolution}, where the improving ability to revise answers can be distilled back into generation performance with regular teacher synchronization.

\centerline{\faGithub~\href{https://github.com/princeton-pli/STAT}{https://github.com/princeton-pli/Self-Distillation-Zero}}

\end{abstract}

\begin{figure}[h]
    \centering
    \vspace{-10pt}
    \captionsetup{font=small}
    \includegraphics[width=0.95\textwidth]{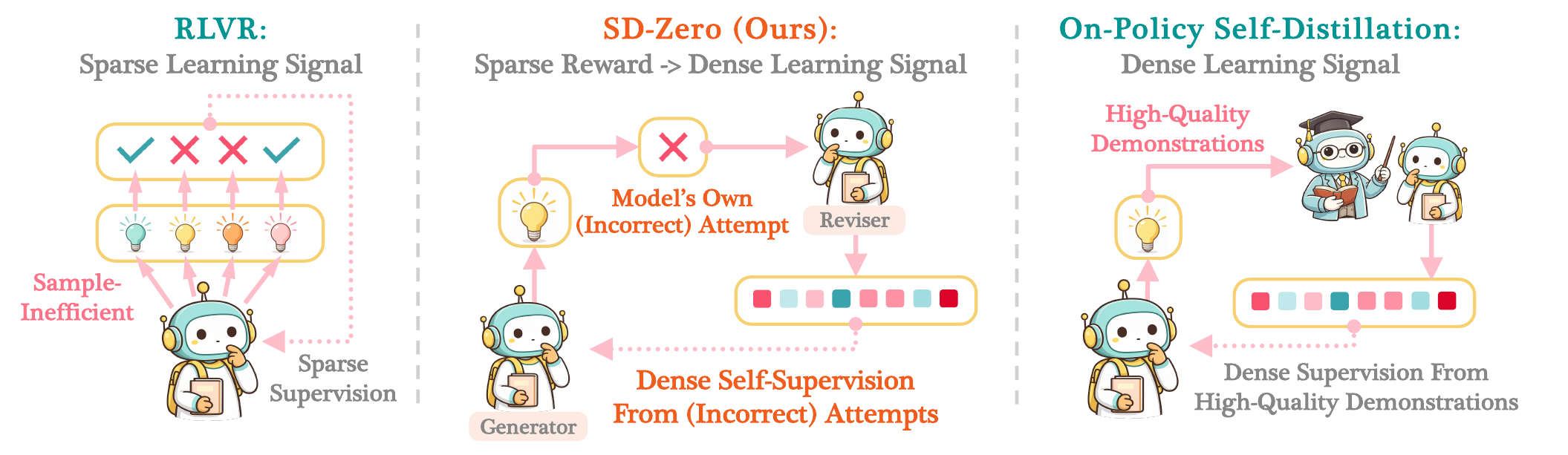}
    \vspace{-5pt}
    \caption{
    % Reinforcement learning from verified rewards (RLVR) provides sparse learning signals; on-policy distillation relies on an external teacher to provide dense supervision. 
    We introduce \textbf{\ours{}}, an on-policy self-distillation paradigm where the teacher (\emph{reviser}) can condition on incorrect student (\emph{generator}) attempts.
    % that 
    % converts sparse rewards into dense self-supervision through self-revision feedback, without relying on high-quality demonstrations.
    % \danqi{This is my biggest confusion. This figure is about how self-distillation is different from RLVR and onpodi, but not about how your method is different from other self-distillation methods. There is no generator or reviser in the figure.}
    }
    \label{fig:starter}
\end{figure}

\section{Introduction}

\looseness-1Reinforcement learning (RL) has become the dominant approach for post-training language models on reasoning tasks in verifiable settings such as math and coding \citep{shao2024deepseekmath,Guo_2025,chen2025minimax,zheng2025group}.
%\danqi{ and coding? cite}
These methods require only a binary reward, typically whether the final answer is correct, that makes them broadly applicable. However, a binary reward per response provides no information about which intermediate reasoning steps were sound. This sparse supervision makes training expensive: the model must discover good reasoning by comparing and contrasting many self-generated responses.

%\yj{This is probably a very minor point, but both OPD and SD are on-policy distillation methods. So maybe it's better could first introduce OPD as a method that usually relies on an external teacher, and then mention that, "more recently, there have been attempts to remove the external teacher. However, it still requires …" (something like this). But this is just a minor taxonomy point.}

\begin{figure}[t]
    \centering
    \captionsetup{font=small}
    \includegraphics[width=\textwidth]{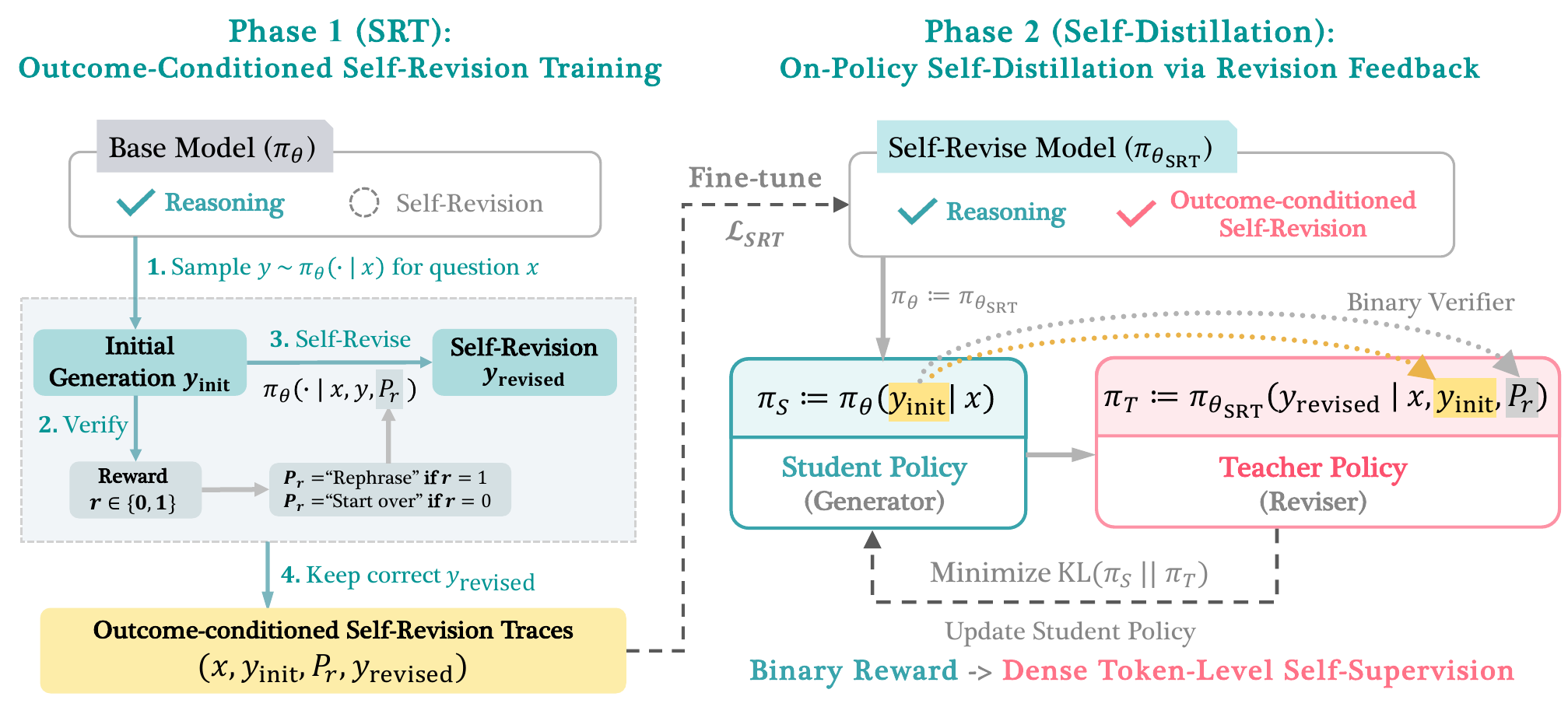}
    \caption{\textbf{Overview of \ours{}.} In \textbf{Phase 1 (\sft{})}, we collect 6K outcome-conditioned self-revision traces by sampling an initial response from the base model, prompting the model to self-revise its incorrect response, and keeping the correct self-revision. In \textbf{Phase 2 (\opsd{})}, we conduct on-policy self-distillation with the self-revise (\sft{}) model acting as both student and teacher: the \textit{student} generates an on-policy response, and the \textit{teacher} generates a revised version conditioned on that response and its outcome reward. Throughout the \opsd{} phase, the model bootstraps performance through internalizing self-revision behavior. See \Cref{alg:pipeline} for details.
    % \liam{I'm very confused by this diagram - and unless I'm being dumb, I feel like the OPSRD loss setup might not make total sense (notationally)?}
    % \yh{make things consistent with section 2}
    % \\\yj{I'm confused of what are these check signs? So to my understanding, regardless of correctness, the model self-revises the response and filters out only the correct revised response? Also, it's hard to get the fine-tuning objective with this "self-revision traces" and that we are "minimizing" this KL. Finally, I think notations are not consistent with the explanation ($z$ and $y_{revisied}$)?}
    % \danqi{What is $\theta^*$? Do generator and reviser always share weights?}
    }
    
    \label{fig:main-figure}
\end{figure}

\looseness-1Distillation methods provide a more sample-efficient alternative by converting supervision into token-level feedback on student-generated responses. \emph{On-policy distillation} methods \citep{agarwal2024onpolicy,gu2024minillm,boizard2024uld,minixhofer2025cross,lu2025onpolicydistillation} assume access to an external stronger teacher that can provide token-level feedback on the student's responses. More recent \emph{self-distillation} methods, including OPSD \citep{zhao2026self}, SDFT \citep{shenfeld2026self}, and SDPO \citep{hubotter2026reinforcement}, remove the external teacher but still require high-quality demonstrations that are much better than the model's responses.
These demonstrations are typically assumed to come from an external teacher (OPSD, SDFT) or repeated generation and filtering from the model itself (SDPO) \footnote{See \Cref{tab:comparison} for a detailed comparison of post-training methods.}. Collecting such supervision can be unavailable or prohibitively expensive to collect. This raises the central question of our work:

\begin{quote}
    Can the model condition its own initial attempts (possibly incorrect) and their sparse rewards, and provide improved dense supervision to itself?
\end{quote}

%\ap{we can even condition on incorrect response}
%\citep{agarwal2024policy,gu2024minillm,boizard2024uld,minixhofer2025cross,lu2025onpolicydistillation,zhao2026self,shenfeld2026self}. On-policy distillation
%At each token position, the student is trained to match the predictions of a teacher. However, these approaches typically require either token-level supervision from a stronger teacher.. 

% \begin{quote}
%     Can we obtain token-level supervision from sparse outcome feedback, without access to a strong teacher or ground-truth solutions?
%     \ap{}
% \end{quote}
 
\looseness-1\textbf{\method{} (\ours{}):}
Our method is built around a single model that plays two roles: a \textit{generator} and a \textit{reviser}. As a generator, the model produces candidate responses to a given question. As a reviser, the same model is conditioned on its own response and reward, and either {corrects} an \emph{incorrect} response or {rephrases} a \emph{correct} one.

%\danqi{why do we need to rephrase a correct one? is it possible the answer is correct but intermediate solution is wrong?} \danqi{Maybe you will make it clearer later but i still feel demonstration / answer are not so well defined and the verifier is only applied to the answer.}

\looseness-1\ours{} proceeds in two phases. In the first phase, we collect self-revision traces by sampling responses, checking for correctness, and prompting the model to revise incorrect ones. We retain only traces where revision succeeds. We fine-tune on these filtered traces, a procedure we call Self-Revision Training (\sft{}).
In the second phase (self-distillation), we use the reviser as a teacher to provide token-level supervision over the generator's responses, thereby transforming outcome-level reward into dense token-level supervision.

% Our method instantiates a single model in two roles: a \textit{generator} and a \textit{reviser}. As a generator, the model produces candidate responses to a given question. As a reviser, the same model is conditioned on its own response and its correctness, and either \textit{revises} an incorrect response or \textit{rephrases} a correct one. \ours{} then uses the reviser as a teacher to provide token-level supervision over the generator’s sampled responses, thereby converting sparse outcome feedback on the responses into dense token-level feedback. To enable this behavior, \ours{} begins with an initial training phase in which an off-the-shelf model is trained to perform both revision and rephrasing on its own sampled responses.\ap{make sure intro is self-contained}

\textbf{Key Findings.} Applying \ours{} on Olmo and Qwen models on math and code reasoning datasets shows the following: 
% \citep{openr1}\danqi{why cite OpenR1 here?} 
\begin{itemize}
    \item \looseness-1\textbf{Phase 1 (\sft{}) alone outperforms all baselines on the same data budget.}
    Training on 6K self-generated revision traces improves average accuracy by 7.8\% for \qwen{} and 9.2\% for \olmo{} (\Cref{tab:main_results}). 
    
    % \item \looseness-1\textbf{Phase 2 (\opsd{}) is essential for the major efficiency gains of \ours{} at both training and inference time.}
    % Models trained only with \sft{} tend to produce excessively long and verbose responses. In contrast, \opsd{} teaches the model to internalize its \textit{revision} behavior, enabling it to generate stronger responses directly as the \textit{generator}. As a result, responses after \opsd{} are approximately $2\times$ shorter than those from \sft{} (\Cref{fig:self-revision}). Beyond this efficiency benefit, \opsd{} also delivers an additional $2.7\%$ gain for \qwen{} and $1.2\%$ for \olmo{} over \sft{}, bringing the total improvement over the base models to $10.5\%$ and $10.4\%$, respectively.
    % \emph{Crucially}, this second phase is what makes \ours{} highly sample-efficient during training: \opsd{} requires only a single response per question, whereas the \sft{} phase must collect multiple sampled responses to construct revision traces. We provide a more detailed analysis in \Cref{sec:distillation-results}.
    
    \item \looseness-1\textbf{Phase 2 (\opsd{}) is critical for the training and inference efficiency of \ours{}.}
     \sft{} trained model produces overly long responses. \opsd{} trains the model to internalize its \textit{revision} behavior and generate stronger answers directly as the \textit{generator}, reducing response length by roughly $2\times$ (\Cref{fig:self-revision}). It also improves performance beyond \sft{}, yielding an additional $2.7\%$ gain for \qwen{} and $1.2\%$ for \olmo{}, for total gains of $10.5\%$ and $10.4\%$ over the base models. \emph{Importantly}, \opsd{} is also what makes \ours{} training sample-efficient during training: it requires only one response per question, whereas \sft{} must sample multiple responses to construct revision traces (\Cref{sec:distillation-results})
    %\opsd{} yields an additional $2.7\%$ gain for \qwen{} and $1.2\%$ for \olmo{} beyond \sft{}, resulting in total gains of $10.5\%$ and $10.4\%$, respectively, over the corresponding base models. 

    % \\\sk{For phase 2 bullet, I think we should lead with the problem phase 1 caused (very long generations) that phase 2 fixes. I'm worried if a lazy reviewer sees the 2.7\% gains first, they might conclude that Phase 2 offers marginal improvement / isn't needed. But if they first read that phase 2 compresses verbose revision into efficient generation, I think the story would read as "phase 1 unlocks the capability, phase 2 makes it practical"}

    \item \looseness-1\textbf{\ours{} enables iterative self-evolution through regular teacher synchronization.}
    Because training also improves the model’s revision capability, the updated model can serve as the reviser teacher in subsequent rounds of \opsd{}. After one epoch of \opsd{}, synchronizing the teacher with the updated model yields a further performance gain of at least $3\%$ (\Cref{fig:evolution}), suggesting that \ours{} can continue to improve over multiple rounds of \opsd{}.
\end{itemize}

\section{\ours{}: Turning Binary Rewards into Dense Self-Supervision} \label{sec:method}

% \danqi{Title: reward or rewards}

%\yj{Also, how about having a clear summary of the full method with bullet points? - Phase1:... - Phase 2: ...}
We consider a dataset $\mathcal{D} = \{(x_i, a_i)\}_{i=1}^N$, where $x_i$ denotes an input problem and $a_i$ its corresponding ground-truth final answer. Crucially, we do not assume access to gold solutions. Let $\pi_\theta$ denote the student policy with parameters $\theta$, which generates a reasoning response $y$ for input $x$ according to $\pi_\theta(y \mid x)$. For each example $(x, a)$ and generated response $y$, we assume access to a binary reward $r(y,a) \in \{0,1\}$, where $r(y,a)=1$ if the final answer extracted from $y$ matches $a$, and $r(y,a)=0$ otherwise.

The core design of \ours{} is around a single model that can play two roles: a \emph{generator} and a \emph{reviser}. We partition the training set into two disjoint subsets of sizes $N_1$ and $N_2$:
\begin{itemize}
\setlength{\itemsep}{-0.2em}
\item  \textbf{Phase 1, Self-Revision Training}: Using the first subset of $N_1$ examples, we train the model to perform both the generator and reviser roles.\newline
\item   \textbf{Phase 2, \opsd{}}: Using the remaining $N_2$ examples, we leverage the reviser to transform sparse outcome-level supervision into dense token-level supervision over the generator's responses.
\end{itemize}

In our experiments, we refer to the model obtained after Phase 1 as the \sft{} model, and the model obtained after Phase 2 as the \ours{} model.

\subsection{Phase 1 (\sft{}): Self-Revision Training} \label{sec:phase1}

\looseness-1Our base models exhibit strong \textit{generator} performance but weak \textit{reviser} performance (\Cref{fig:self-revision}). This phase strengthens the model as a \textit{reviser}, while also improving it as a \textit{generator}.

\paragraph{Outcome-Conditioned Self-Revision.}
For each input $x$, we sample multiple initial reasoning responses $y_\text{init} \sim \pi_\theta(\cdot \mid x)$. Given a sampled response $y_\text{init}$, we construct a revision prompt that conditions on the original problem, the sampled response, and a control phrase indicating whether the response should be revised or simply rephrased: 
\[
P_{r} =
\begin{cases}
\text{``Let me rephrase the above solution.''}, & \text{if } r(y_\text{init},a)=1, \\
\text{``Wait, this response is not correct, let me start over.''}, & \text{if }  r(y_\text{init},a)=0.
\end{cases}
\]
We then use the same model to generate a revised response:
\[
y_\text{revised} \sim \pi_\theta(\cdot \mid x, y_\text{init}, P_{r}).
\]

For correct initial responses, this encourages the model to produce a rephrase of a correct solution, for incorrect initial responses, this encourages the model to critique on the previous attempt and regenerate a correct solution. We observe that when asked to rephrase a correct response, the model typically produces a shorter answer, which may additionally encourage more concise responses during the second phase. We defer further discussion to \Cref{sec:experiments}. We create a dataset $\mathcal{D}_{\textsc{revision}} =  \{(x,y_\text{init},P_r, y_\text{revised})\}$.

% \[
% \Tilde{y} \sim \pi_\theta(\cdot \mid x, y, P_{r}).
% \]
%\yj{Again, $z$ or $y_{revised}$?} 
%\sk{I like $y_{revised}$ a bit better, I think it better communicates the meaning}
% \yj{Also, "Rephrase" needs more justification as it introduces extra compute cost and the original y is also sampled from the student model}\jiarui{+1 I feel like we need to explain why we do rephrasing}
 
% \paragraph{Self-revision dataset construction.}
% We create two datasets as follows:
% \begin{align*}
% \mathcal{D}_{\textsc{revision}} &= \{(x,y,P_r, \Tilde{y}) \,:\, r(\Tilde{y}, a)=1, r(A, a) = 0\}, \\
% \mathcal{D}_{\textsc{rephrase}} &= \{(x,y,P_r, \Tilde{y}) \,:\, r(\Tilde{y}, a)=1, r(A, a) = 1\}
% \end{align*}
% \yh{hide notation for D-rephrase}

%, which has the following two components:

%\textbf{Self-revision objective.}
% We then train the student on the generated datasets with loss $\mathcal{L}_{\sft{}}(\theta)$: 
%Now, we train the model with the following loss $\mathcal{L}_{\text{\sft{}}}$ consisting of two log-likelihood terms $\mathcal{L}_{\text{generation}}$ and $\mathcal{L}_{\text{revision}}$:
% on $(y_\text{init},P_r, y_\text{revised})$, conditioned on the input $x$ in the dataset $D_{\textsc{revision}}$: 
%on a combination of two objectives:

% yj: I think this looks a bit visually dense right now. It may read better if the revision loss, generation loss, and final objective are written as separate numbered equations rather than in a single small-font aligned block. That would make each term easier to parse and easier to reference later in the paper. => Made an edit
\textbf{Self-Revision Objective.} We train the model on two tasks simultaneously: given the input, initial attempt, and reward signal, produce a corrected or rephrased response (revision); and given only the input, produce the full correct response from scratch (generation). The corresponding loss $\mathcal{L}_{\text{\sft{}}}$ consists of two log-likelihood terms $\mathcal{L}_{\text{revision}}$ and $\mathcal{L}_{\text{generation}}$.

The \emph{revision loss} trains the model to produce $y_\text{revised}$ when conditioned on the input $x$, the first attempt $y_\text{init}$, and the reward prompt $P_r$:
{\small
\begin{equation}
\mathcal{L}_{\text{revision}}(\theta)=
\mathbb{E}_{(x,y_\text{init},P_r, y_\text{revised})\sim \mathcal{D}_{\textsc{revision}}}
\left[
-\sum_{t=1}^{|y'|}
\log \pi_\theta\!\left(y'_t \mid x, y_\text{init}, P_r, y'_{<t}\right)
\right],\text{where } y' = y_\text{revised}.
\label{eq:revision}
\end{equation}
}
The \emph{generation loss} retains the model's ability to produce the full correct response conditioned only on input $x$:
{
% \small
\fontsize{8.8pt}{9pt}\selectfont
\begin{equation}
\mathcal{L}_{\text{generation}}(\theta)=
\mathbb{E}_{(x,y_\text{init},P_r, y_\text{revised})\sim \mathcal{D}_{\textsc{revision}}}
\left[
-\sum_{t=1}^{|y'|}
\log \pi_\theta\!\left(y'_t \mid x, y'_{<t}\right)
\right],\text{where } y' = [y_\text{init},P_r, y_\text{revised}].
\label{eq:generation}
\end{equation}
}
The overall self-revision objective combines these two terms:
{
\begin{equation}
\mathcal{L}_{\text{\sft{}}}(\theta)=
\mathcal{L}_{\text{revision}}(\theta) + \mathcal{L}_{\text{generation}}(\theta).
\label{eq:sft}
\end{equation}}

\looseness-1 $\mathcal{L}_{\text{revision}}$ (Eq.~\ref{eq:revision}) trains the model to self-revise, i.e., producing $y_\text{revised}$ conditioned on the input $x$, the first attempt $y_\text{init}$, and the prompt $P_r$, while $\mathcal{L}_{\text{generation}}$ (Eq.~\ref{eq:generation}) retains the model's generation capability, i.e., producing correct response $y$ conditioned on input $x$. This combination implicitly teaches the model to actively evaluate its current response at inference time and generate a self-revision if necessary. As a result, the trained model often exhibits explicit self-revision behaviors that lead to extremely long responses (see \Cref{sec:srt-results,sec:distillation-results}). In the next phase, we perform on-policy distillation on the \sft{} model, distilling this learned revision behavior back into the generator to enable sample-efficient self-improvement.

\subsection{Phase 2 (\opsd{}): On-Policy Self-Distillation via Revision Feedback}\label{sec:phase2}

%\yh{put motivation here}
\opsd{} phase distills the reviser’s behavior back into the generator, so that the model can internalize revision and produce more compact and well-directed responses. Concretely, we leverage the self-revision capability learned in Phase 1 to perform on-policy self-distillation. Let $\theta_{\sft{}}$ denote the model obtained at the end of Phase 1. In Phase 2, we initialize the student parameters as $\theta:=\theta_{\sft{}}$. 
The \textbf{generator} (student) uses the current parameters $\theta$ to produce on-policy responses $\pi_{\theta}(\cdot \mid x)$; its parameters are updated throughout Phase 2. The \textbf{reviser} (teacher) remains frozen at $\theta_{\sft{}}$ and provides token-level supervision by conditioning on the generator's response and its binary reward via $\pi_{\theta_{\sft{}}}(\cdot \mid x, y, P_r)$. The generator is trained to match the reviser's distribution with:
%$\theta:=\theta_{\sft{}}$ as the initial model checkpoint that plays both a generator and a reviser role: the generator role serves as the \textit{student} that outputs on-policy responses, while the reviser role serves as the \textit{teacher} that provides token-level supervision over its own responses. The resulting training objective is:
%\yh{Reverse the KL here: KL($\pi_s$ | $\pi_T$)}
% \sk{in the code, are you overriding $\alpha=0$ with $\alpha=1$? my understanding is that the sdft continual learning default code sets $\alpha=0$  (forward KL)}
% \begin{align*}
% \mathcal{L}_{\text{\opsd{}}}(\theta)
% &=
% \mathbb{E}_{(x,a)\sim\mathcal{D}}
% \;
% \mathbb{E}_{y\sim\pi_\theta(\cdot\mid x)}
% \sum_{t=1}^{|y|}
% \KL\!\left(
% \pi_\theta(\cdot \mid x, y_{<t})\;\|\;
% \pi_{\theta_{\sft{}}}(\cdot \mid x, y, P_{r}, y_{<t})\right).
% % \\
% % &\approx \mathbb{E}_{(x,a)\sim\mathcal{D}}
% % \;
% % \mathbb{E}_{y\sim\pi_\theta(\cdot\mid x)}
% % \sum_{t=1}^{|y|} \log \pi_\theta(y_t \mid x, y_{<t}) - \log \pi_{\theta_{\sft{}}}(\cdot \mid x, y, P_{r}, y_{<t})
% \end{align*}
{\small
\begin{equation}
\mathcal{L}_{\text{\opsd{}}}(\theta)
=
\mathbb{E}_{(x,a)\sim\mathcal{D}}
\;
\mathbb{E}_{y\sim\pi_\theta(\cdot\mid x)}
\sum_{t=1}^{|y|}
\KL\!\left(
\pi_\theta(\cdot \mid x, y_{<t})\;\|\;
\pi_{\theta_{\sft{}}}(\cdot \mid x, y, P_{r}, y_{<t})\right).
\label{eq:opsd}
\end{equation}
}
%\yh{put a simplified version of algorithm here}\\
% \sk{``in words'' alg summary below:} \\

\looseness-1 For each training example, a response is sampled from the model and checked against the ground-truth final answer for correctness. The Phase 1 reviser, conditioned on the model's response and whether it was correct, produces a next-token distribution at each token position. The model is trained to match this distribution via KL divergence loss.

\section{Experiments}
\label{sec:experiments}

%\sk{Potential restructure for section 3+4 (combined) below; relevant comments from section 3+4 still left inline} \\
% \sk{below is newer exp + discussion sections; relevant comments from old version kept inline; to use old experiment + discussion sections, uncomment in colm2026\_conference.tex file}

\def\changesize#1{\fontsize{#1}{10.5pt}\selectfont}
\begin{table*}[t]
\centering
\small
\fontsize{8.7pt}{10.5pt}\selectfont
    \renewcommand{\arraystretch}{1.18}
    \setlength{\tabcolsep}{2.7pt}
% \begin{tabular}{l@{\hspace{1.8pt}}c@{\hspace{2.3pt}}c@{\hspace{3pt}}c@{\hspace{2.3pt}}c@{\hspace{3pt}}c@{\hspace{2.3pt}}cc@{\hspace{2.3pt}}cc}
\begin{tabular}{l@{\hspace{1.6pt}}ccc@{\hspace{2pt}}c@{\hspace{2pt}}ccccc}
\toprule
 & \multicolumn{6}{c}{\textbf{Math}} & \multicolumn{2}{c}{\textbf{Code}} & \multirow{2}{*}{\textbf{Avg.}} \\
\cmidrule(lr){2-7} \cmidrule(lr){8-9}
 & {\changesize{8.3} AIME24} & {\changesize{8.3}AIME25} & {\changesize{8.3}HMMT25} & {\changesize{8.3}AMOBench} & {\changesize{8.3}OpenR1} & {\changesize{8.3}MATH} & {\changesize{8.3}Codeforces} & {\changesize{8.3}LCB} & \\
\midrule
\textbf{\qwen{}}           & 59.6 & 45.8 & 26.7 &  9.8 & 55.8 & 91.0 & 48.0 & 61.8 & 49.8 \\
SFT                        & 61.7 & 46.3 & 31.3 &  7.3 & 56.1 & 91.3 & 49.1 & 57.2 & 50.0 \\
RFT                        & 64.2 & 52.1 & 37.1 & \underline{11.3} & 59.3 & 91.9 & 50.9 & 68.0 & 54.3 \\
GRPO                       & 62.5 & 50.0 & 30.4 & 11.0 & \textbf{62.9} & \underline{93.5} & 52.2 & 62.6 & 53.1 \\
SDFT                       & 63.3 & 47.9 & 32.9 &  9.0 & 57.0 & 91.1 & 49.3 & 59.2 & 51.2 \\

% \rowcolor{lightyellow!50}
\textbf{\sft{}} (Ours)     & \underline{66.7} & \underline{59.2} & \underline{40.0} & \textbf{16.0} & 59.8 & 92.4 & \underline{52.7} & \underline{74.4} & \underline{57.6} \\

\rowcolor{lightblue!80}
\textbf{\ours{}} (Ours)    & \textbf{68.3} & \textbf{60.0} & \textbf{45.4} & \textbf{16.0} & \underline{60.4} & \textbf{93.6} & \textbf{56.1} & \textbf{82.6} & \textbf{60.3} \\
\midrule

\textbf{\olmo{}}           & 56.7            & 42.1            & 25.0            & 1.3            & 48.9            & 91.1            & 31.7            & 32.4            & 41.1            \\
SFT                        & 51.3            & 43.8            & 30.4            & 1.3            & 48.5            & 91.4            & 31.7            & 41.0            & 42.4            \\
RFT                        & 56.7            & 48.3            & 35.8            & 2.3            & 50.9            & 91.4            & 39.2            & 49.4            & 46.7            \\
GRPO                       & 54.6            & 43.8            & 25.8            & \underline{4.8} & \textbf{56.9}   & 91.8            & 37.1            & 43.6            & 44.8            \\
SDFT                       & 52.9            & 45.0            & 32.1            & 1.3            & 49.0            & 91.2            & 33.5            & 42.3            & 43.4            \\
\textbf{\sft{}} (Ours)     & \underline{59.2} & \underline{52.9} & \underline{39.6} & 3.5            & 52.4            & \underline{92.3} & \underline{42.8} & \textbf{59.6}   & \underline{50.3} \\
\rowcolor{lightblue!80}
\textbf{\ours{}} (Ours)    & \textbf{61.7}   & \textbf{53.8}   & \textbf{40.4}   & \textbf{5.5}   & \underline{55.3} & \textbf{94.0}   & \textbf{43.5}   & \underline{57.8} & \textbf{51.5}   \\
\bottomrule
\end{tabular}
\caption{\textbf{Performance comparison of \ours{} and \sft{} (Self-Revision Training, Phase~1) against baseline post-training methods} on math and code reasoning benchmarks, reported as avg@8. Across both \qwen{} and \olmo{}, \sft{} and especially \ours{} achieve the strongest overall results; “LCB” denotes LiveCodeBench. All methods are compute equalized (see Appendix \ref{app:compare-budget} for more details).}
\vspace{-10pt}
\label{tab:main_results}
\end{table*}

\subsection{Experimental Setup}
\label{sec:setup}

\looseness-1\textbf{Models.} We use \qwen{} \citep{yang2025qwen3technicalreport} and \olmo{} \citep{olmo2025olmo3} as base models. All sampling uses temperature 0.7 with a 16K token limit during training and a 32K limit at evaluation. We report avg@8 throughout.

\looseness-1\textbf{Datasets.} We train on two domains separately: (1)~\textbf{OpenR1-Math} \citep{openr1}, from which we select 15K competition- and olympiad-level problems with verified solutions, and (2)~\textbf{Codeforces} \citep{penedo2025codeforces}, from which we select 7.5K samples from the cpp (\texttt{solutions}) subset and 7.5K from the Python (\texttt{solutions\_py}) subset. We include more details on training data curation in Appendix \ref{app:training-data}.

% \sk{also include sampling budget + more details for revision sper question num questions, etc}

\looseness-1\textbf{Evaluation. } We evaluate on eight benchmarks in math and code domains: competition math (AIME24 \citep{aime24}, AIME25 \citep{aime25}, HMMT25 \citep{hmmt_2025}, MATH \citep{hendrycks2021measuringmathematicalproblemsolving}), olympiad math (AMOBench~\citep{an2025amobenchlargelanguagemodels} and OpenR1-Math \citep{openr1}), and competitive programming (Codeforces \citep{penedo2025codeforces} and LiveCodeBench \citep{jain2024livecodebenchholisticcontaminationfree}). For the two in-distribution datasets we hold out 500 test questions each.

\looseness-1All baselines train on the same 15K questions: (1)~\textbf{SFT} on high-quality demonstrations from DeepSeek-R1~\citep{Guo_2025}, (2)~\textbf{RFT}, rejection fine-tuning on correct self-generated traces \citep{yuan2023scalingrelationshiplearningmathematical}, (3)~\textbf{GRPO} with binary correctness reward \citep{grpo} \footnote{\looseness-1We use DAPO (\citet{yu2025dapo}), an improved and commonly used variant of GRPO.}, and (4)~\textbf{SDFT}, on-policy self-distillation with the model conditioned on high-quality demonstrations as its own teacher \citep{zhao2026selfdistilledreasoneronpolicyselfdistillation, shenfeld2026selfdistillationenablescontinuallearning}. 
% \danqi{Why do you call it OPSD, not SDFT? Does it always have access to CoT from DeepSeek-R1? You probably need a baseline that the teacher has access to only input and answer?}
We attach a detailed comparison of sampling budgets across \ours{} and baseline methods in Appendix \ref{app:compare-budget}, and training hyperparameters in Appendix \ref{app:hyperparameters}.

%\subsection{Self-Revision Training Performs Stronger than Training on Expert Solutions or Filtered Self-Generations}
\subsection{SRT Outperforms Training on Expert Solutions or Filtered Self-Generations}
%\subsection{\ours{} improves the }
\label{sec:srt-results}

%\textbf{Self-Revision Training Teaches Inference-Time Error Correction Better.} 
%\ap{take a pass; maybe shorten}
%One might expect that expert solutions (SFT) or filtered correct self-generations (RFT) would provide a strong offline training signal. 

\looseness-1We first compare \sft{} against training on high-quality demonstrations (SFT) and filtered correct self-generations (RFT) (Table~\ref{tab:main_results}). Trained on only 6K self-revision responses, \sft{} improves average accuracy by 7.8\% on \qwen{} and 9.2\% on \olmo{}, substantially outperforming both SFT and RFT trained on 15K examples. The baselines fail for different reasons: SFT on DeepSeek-R1 responses degrades \qwen{} on benchmarks such as AMOBench and LiveCodeBench, while RFT shows minimal gains on harder tasks such as AMOBench.

\looseness-1\sft{} is effective because it supervises error correction: each example pairs an on-policy incorrect attempt with a self-corrected revision; we find that correctness filtering of revision traces is important (Appendix \ref{app:countdown}). Unlike RFT, which removes incorrect reasoning entirely, \sft{} preserves the failed attempt as context, allowing the model to learn from its own mistakes. This structure yields especially strong gains on harder benchmarks such as AIME25, HMMT25, and LiveCodeBench.

% \looseness-1We first compare performance of \sft{} with training on an high-quality demonstrations (SFT) or filtered correct self-generations (RFT) (Table~\ref{tab:main_results}). \sft{}, trained on only 6K self-revision responses, improves average accuracy by 7.8\% on \qwen{} and 9.2\% on \olmo{}, significantly surpassing both SFT and RFT training on 15K examples. 
% The two baselines each fail for separate reasons: SFT on DeepSeek-R1 responses leads to degradation in the performance of \qwen{} on benchmarks such as AMOBench and LiveCodeBench. RFT  shows minimal improvement on hard tasks such as AMOBench.

% \looseness-1\sft{}’s effectiveness comes from supervising the process of correcting an incorrect response, with instances containing an on-policy incorrect attempt paired with a self-corrected attempt; we find that correctness filtering of revision traces is important (Appendix \ref{app:countdown}). While SRT retains the incorrect attempt as context, RFT filters out all incorrect reasoning, so the model never sees its own failure.
% This error-correction structure helps \sft{} achieve substantial gains on harder benchmarks such as AIME25, HMMT25, and LiveCodeBench.

% but stalls where the base model rarely solves problems: AMOBench gains only 1.5\% (\qwen{}) and 1.0\% (\olmo{}) under RFT, versus 6.2\% and 2.2\% under \sft{}. 
% \sk{might need more details since we say baselines are trained on same 15k examples and experiments are compute controlled}
% (AIME25 +13.4\%, HMMT25 +13.3\%, LiveCodeBench +12.6\% for \qwen{})
%, despite training only on OpenR1-Math and Codeforces.

\begin{wrapfigure}{r}{0.35\textwidth}
    \vspace{-13pt}
  \centering
  \captionsetup{font=footnotesize}
  \includegraphics[
    width=0.35\textwidth,
    trim=0.7cm 0.1cm 0.1cm 0.4cm,
    clip
  ]{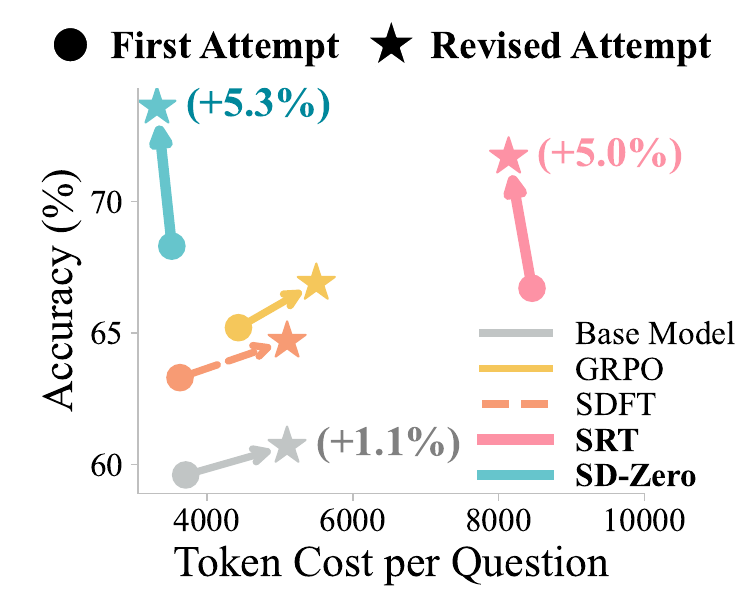}
  \vspace{-15pt}
  \caption{\textbf{Comparison of outcome-conditioned self-revision capability} on AIME24, \qwen{}. \sft{} unlocks self-revision behaviors, while final \ours{} model preserves this advantage while improving token efficiency.}
  \label{fig:self-revision}
  \vspace{-5pt}
\end{wrapfigure}
\looseness-1\textbf{\sft{} Significantly Boosts Self-Revision Capability.}
To measure directly, we run a Generate-then-Revise evaluation on 1K AIME24 questions using \qwen{}: (1) \textit{First Attempt}: sample one response per question, (2) \textit{Revised Attempt}: prompt the model to generate a revision conditioned on the first attempt and its binary correctness reward. As shown in Figure~\ref{fig:self-revision} (detailed statistics in \Cref{tab:revision-token-tradeoff}), while the base model gains only $1.1\%$ from revision, \sft{} gains $+5.0\%$, indicating that the \sft{} model has learned to use outcome reward to correct its first attempt rather than simply resampling a new attempt. 
Notably, the revision responses produced by \sft{} are, on average, even shorter than the model’s first attempts, suggesting more targeted and proactive reasoning during revision; in contrast, revisions from the base model tend to be longer and likely more redundant. More interestingly, the self-revision capability of the model improves further after the \opsd{} phase. This observation is central to the self-evolving nature of \ours{}, which we discuss further in \Cref{sec:self-evolution}.

%This self-revision capability emerges in the \sft{} phase and is preserved through \ours{}, which further improves overall performance with less token cost.
% With \sft{} phase explicitly unlocking self-revision behaviors, the \ours{} phase preserves this gain while further improving overall performance, showing that this learned revision behavior can be transferred to later training stages.
% , using fewer tokens for the revisions than for the first attempt. This suggests that \sft{}
% The contrast is sharp: baselines produce longer, apparently redundant revisions, while the \sft{} model makes targeted corrections.

\subsection{Self-Distillation Distills Revision into Token-Efficient Generation}
\label{sec:distillation-results}

% After \sft{}, the model tends to produce two attempts within a single response, the second revising the first. This is effective but expensive. 
\looseness-1After \sft{}, the model as a generator tends to produce longer responses (as shown in \Cref{fig:self-revision}) with significantly more self-revision behaviors.
\opsd{} phase distills the self-revision behavior of the model back into more proactive responses of the generator.
%Concretely, the student generates a response $y_\text{init}$ given question $x$. The teacher sees $(x, y_\text{init}, P_r)$, the same response together with the binary reward of that attempt, and provides token-level supervision via KL divergence.

\looseness-1\textbf{\opsd{} Enables Stronger, Token-Efficient Generations.}
\Cref{tab:main_results} shows that \opsd{} phase adds $2.7\%$ on \qwen{} and $1.2\%$ on \olmo{} beyond \sft{}, for total gains of $+10.5\%$ and $+10.4\%$ over the base models. For \qwen{}, the gains concentrate on code benchmarks and HMMT25. For \olmo{}, the \opsd{} phase mainly improves on math benchmarks such as AIME24 and AMOBench. We also report pass@8 results in \Cref{tab:pass8_results}. Interestingly, pass@8 improves substantially as well, suggesting that \ours{} does not merely sharpen the output distribution, a behavior often associated with RL-based methods such as GRPO \citep{yue2025does}.

%Pass@8 performance (see \Cref{tab:pass8_results}) also shows a consistent improvement, which suggests that our method is not merely sharpening its output distribution, a phenomenon generally observed with RL based algorithms

%, indicating that our method is 
%not merely sharpening the output distribution, but genuinely guiding the model’s exploration to be more well-directed.
% LiveCodeBench actually drops 1.8\%. 
% Phase~2 appears more helpful for code than for math.\yh{benchmark gains}

\looseness-1Furthermore, as shown in \Cref{fig:self-revision},  \ours{} model generates only half as many tokens as \sft{} trained model and fewer tokens than all baselines, while achieving the strongest overall performance. We provide a deeper analysis  in \Cref{sec:behavioral_analysis}, where we show that distilling from the reviser helps the generator to produce more proactive and efficient responses.
% As such, we can view \ours{} as a process that distills the self-revision of the model into improved model's generation attempts, while actively reducing token cost.

% Its revised attempt requires even fewer tokens, while continuing to gain $5.3\%$.

\looseness-1\textbf{Comparison to \sft{}.} Since most of the performance gains in \ours{} arise during the \sft{} phase, one may question the importance of \opsd{}. We argue that \opsd{} is nevertheless a critical component of the \ours{} pipeline for two reasons. First, \opsd{} reduces response length at inference by at least $2\times$, substantially improving inference-time efficiency (\Cref{fig:self-revision}). Second, \opsd{} is highly sample-efficient during training. Constructing the \sft{} dataset requires collecting multiple sampled responses per question in order to obtain successful revisions, whereas \opsd{} requires only a single response per question and uses the reviser’s token-level feedback to provide dense supervision. We provide a detailed sample budget analysis in Appendix \ref{app:compare-budget}.

%\sk{prev sentence incomplete?}
%\yh{1.less token,2.sample efficient (mention rollout budget)} 

\looseness-1\textbf{Comparison to GRPO and SDFT.}  \ours{} outperforms by at least $4.8\%$ on average across benchmarks. These comparisons are made under a matched generation budget, where all methods use the same set of questions and are normalized by the number of model-generated responses in their respective pipelines.

% \begin{wraptable}{r}{0.34\textwidth}
% \vspace{-10pt}
% \centering
% \def\arraystretch{1.1}
% \setlength{\tabcolsep}{4pt}
% \fontsize{8pt}{8.5pt}\selectfont
% \begin{tabular}{lc}
% \toprule
% Method & \textbf{Avg.} \\
% \midrule
% Base model                & 48.1 \\
% $+$SDFT                      & 49.7 \\
% $+$SDFT (final answers only) & 49.5 \\
% $+$\ours{}            & \textbf{57.3} \\
% \bottomrule
% \end{tabular}
% \captionsetup{font=footnotesize}
% \caption{}
% \label{tab:sdft_final_answers_avg}
% \vspace{-0.8em}
% \end{wraptable}

\looseness-1We emphasize that this improvement is substantial for two reasons. First, unlike SDFT, that require gold solutions for each problem, \ours{} requires only a scalar reward on the model’s first attempt. In \Cref{tab:sdft_final_answers}, we show that SDFT performs much worse when given access only to the final answer but not the gold solutions. 
Second, unlike GRPO, whose training requires a group of sampled responses per question, the \opsd{} phase of \ours{} uses only a single response per question, substantially reducing generation cost.
While we note that GRPO may benefit from additional training epochs, our comparison is under a matched single-epoch budget, where \ours{} achieves stronger performance with comparable total sample cost (see Appendix \ref{app:compare-budget} for detailed budget analysis).

%\textbf{More concretely, \ours{} returns the best yield in performance per the number of generated responses from a given model, without requiring excess to an external teacher.} 

\begin{tcolorbox}[
  colback=lightblue!50,
  colframe=vividblue,
  boxrule=0.8pt,
  arc=2pt,
  left=6pt,
  right=6pt,
  top=4pt,
  bottom=4pt
]
\looseness-1 \textbf{Takeaway:} \ours{} offers the best performance per model generations among the compared methods, while requiring neither an external teacher nor high-quality demonstrations.
\end{tcolorbox}

\begin{figure}[t]
    \centering
    % \begin{subfigure}[t]{0.25\textwidth}
    %     \centering
    %     \includegraphics[width=\linewidth]{figures/token_rewards_correct_vs_incorrect.pdf}
    %     \captionsetup{width=0.9\linewidth}
    %     \caption{}
    %     \label{fig:credit-assignment-a}
    % \end{subfigure}
    % \hfill
    % \begin{subfigure}[t]{0.73\textwidth}
    %     \centering
    %     \includegraphics[
    %         width=\linewidth,
    %         trim=0.5cm 2cm 19cm 0cm,
    %         clip
    %     ]{figures/sample_9013_excerpt_token_reward_heatmap.pdf}
    %     \captionsetup{width=0.9\linewidth}
    %     \caption{}
    %     \label{fig:credit-assignment-b}
    % \end{subfigure}
    \captionsetup{font=small}
    \includegraphics[width=0.25\textwidth]{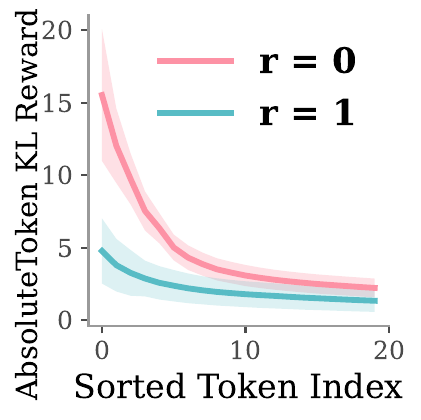}
    \includegraphics[
            width=0.73\textwidth,
            trim=0.5cm 2cm 19cm 0cm,
            clip
        ]{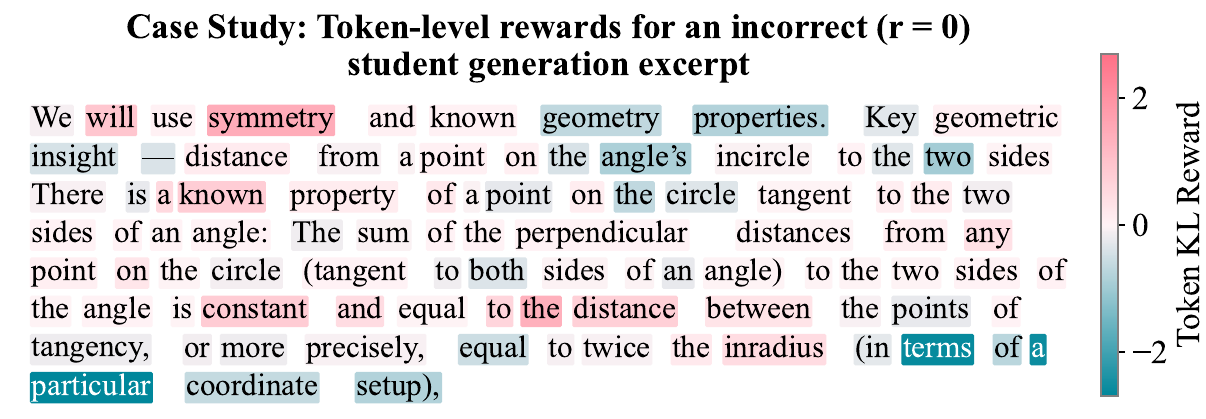}
    \caption{\looseness-1\textbf{Reviser converts binary outcome reward into dense token-level reward.}
\textbf{Left:} Comparison of token-level KL reward distributions for correct (\(\mathbf{r=1}\)) and incorrect (\(\mathbf{r=0}\)) student generations.
Incorrect trajectories concentrate larger rewards on a small number of tokens, whereas correct trajectories receive a flatter reward distribution that mainly preserves the response.
\textbf{Right:} Visualization of token-level KL rewards for an incorrect student generation (\(\mathbf{r=0}\)).
The teacher policy converts binary reward \(r=0\) to dense rewards that localize mistake-relevant tokens and guide targeted correction. 
}
    \vspace{-10pt}
    \label{fig:credit-assignment}
\end{figure}

\subsection{Iterative Self-Evolution: Teacher Synchronization Enables Continued Gains}
\label{sec:self-evolution}

\begin{wrapfigure}{r}{0.34\textwidth}
  \vspace{-10pt}
  \centering
  \captionsetup{font=footnotesize}
  \includegraphics[
    width=0.32\textwidth,
    trim=0.2cm 0cm 0.3cm 0.3cm,
    clip
  ]{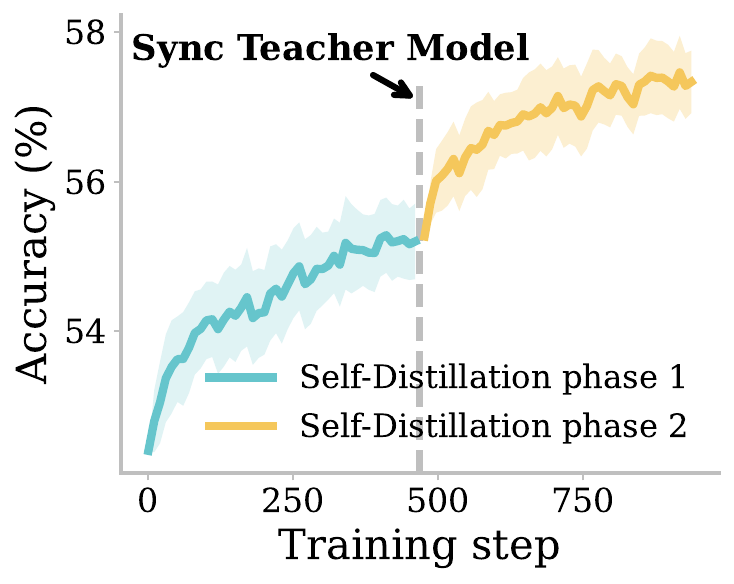}
  \vspace{-5pt}
  \caption{\textbf{Self-evolved reasoning through teacher synchronization in \ours{}.} \ours{} can iteratively improve the model by reusing its own learned self-revision behavior as supervision.}
  \label{fig:evolution}
  \vspace{-20pt}
\end{wrapfigure}

\looseness-1In \ours{} setup, the teacher is fixed as the \sft{} model throughout Phase~2. Performance eventually plateaus because the supervision signal is bounded by a stale teacher. But \opsd{} phase also improves the model's revision capability: the \ours{} model achieves a 5.3\% revision gain on the Generate-then-Revise evaluation, comparable to \sft{}'s 5.0\% (\Cref{fig:self-revision}). 
The improved student can therefore serve as a stronger teacher, enabling what we call \textbf{\emph{iterative self-evolution}}. The capability to revise answers is distilled back into generation, and regular teacher synchronization allows the loop to sustain.

\looseness-1We test this by synchronizing the teacher with the trained student after one epoch and continuing training. Figure~\ref{fig:evolution} shows the result on OpenR1-Math (\qwen{}): the first \opsd{} phase saturates around step 400; after teacher synchronization, a second phase yields at least 3 additional percentage points without signs of saturation. Once primed by \sft{}, the model can continue to self-evolve through iterated teacher synchronization, requiring only the initial attempt and corresponding binary reward in the teacher's context.

% \section{Deeper Analysis of \ours{}}
\section{Understanding How \ours{} Improves Reasoning}
% \label{sec:ablations}

%\vspace{-1em}
% \yh{restructure}
% Here, we provide some analysis into the behavior of  different design choices in \ours{}.
% \yh{intro}

\looseness-1To understand why \ours{} improves reasoning, we provide three analyses: \Cref{sec:credit-assignment} shows that the reviser gives highly localized token-level feedback; \Cref{sec:behavioral_analysis} shows that the model gradually learns shorter, more concise reasoning during \ours{}; and \Cref{sec:ablations} validates key design choices in \ours{}.

\subsection{Token-Level Self-Localization: The Reviser Identifies Which Tokens to Correct}
\label{sec:credit-assignment}

% \sk{Since the abstract discusses (a) token-level self-localization and (b) iterative self-evolution together, would it make sense to keep (a) and (b) in the same section in the main paper? (Currently (a) is in ablations in Section 4, but (b) is in Section 3)}

\looseness-1Although the reviser receives only a binary outcome \(r \in \{0,1\}\), its feedback concentrates on a small subset of tokens, a property we call \emph{token-level self-localization}. We decompose the \opsd{} loss into token-wise terms
\[
D_{\mathrm{KL}}^{(t)} \;:=\; 
D_{\mathrm{KL}}\!\bigl(\pi_\theta(\cdot \mid x, y_{<t}) \;\big\|\; \pi_{\theta_{\sft{}}}(\cdot \mid x, y, P_r, y_{<t})\bigr),
\]
\looseness-1which is approximately the log-probability gap between the generator and reviser at token \(t\). We define \textbf{Token KL Reward} at token \(t\) as: $\log \pi_\theta(y_t \mid x, y_{<t}) - \log \pi_{\theta_{\sft{}}}(y_t \mid x, y, P_r, y_{<t})$.

\looseness-1To quantify how this signal is distributed, we sort tokens within each response by \(D_{\mathrm{KL}}^{(t)}\), divide them into \(20\) equal-sized buckets, and average within each bucket across $200$ responses, separately for correct (\(r=1\)) and incorrect (\(r=0\)) generations. As shown in Figure~\ref{fig:credit-assignment}, the distributions differ sharply: for incorrect responses, most of the KL mass is concentrated on a small fraction of tokens, whereas for correct responses it is much more uniform.

\looseness-1Figure~\ref{fig:credit-assignment} (right) illustrates this on one incorrect generation. Tokens corresponding to the faulty symmetry-based argument receive large positive KL, while tokens associated with the correct coordinate-based solution receive large negative KL. Thus, the reviser does more than penalize incorrect reasoning: it both localizes the error and redirects the model toward a better alternative, converting a scalar outcome into a two-sided token-level signal.

\begin{figure}[t]
    \centering
    \includegraphics[width=0.57\textwidth]{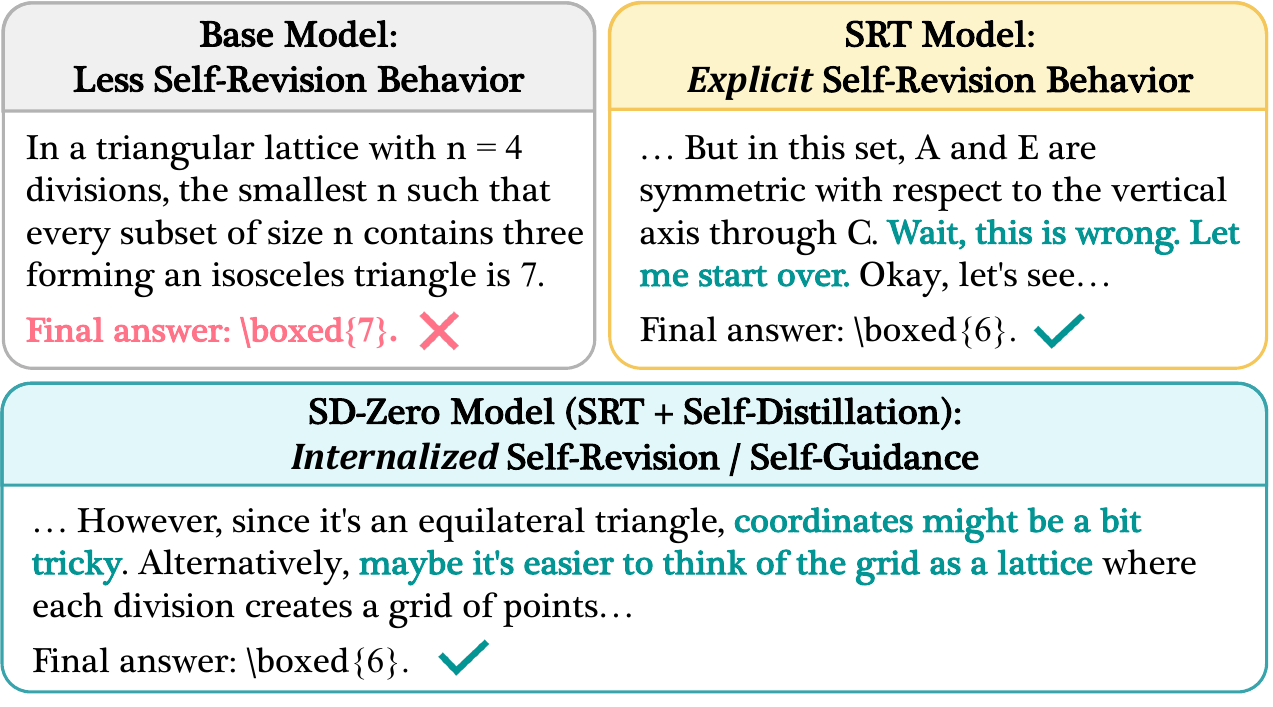}
    \includegraphics[width=0.42\textwidth,
    trim=0cm 0.1cm 0.1cm 0.4cm,
    clip
    ]{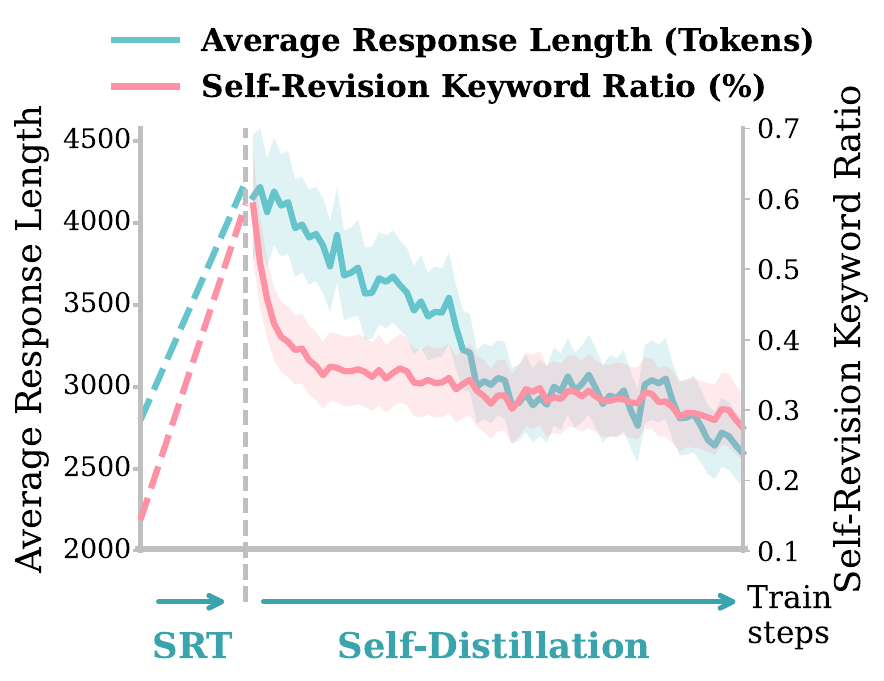}
    \vspace{-1em}
    \captionsetup{font=small}
    \caption{\textbf{Evolution of self-revision behavior across training phases} for \qwen{} on \openmath{}. \textbf{Left:} Example reasoning traces from different model phases. The base model fails without revision, whereas the \sft{} model performs explicit self-revision and the \ours{} model exhibits more internalized self-guidance, identifying pitfalls and directing itself to the correct answer. \textbf{Right:} Training dynamics of self-revision behavior. Response length and explicit self-revision keywords rise during the SRT phase and fall during the \opsd{} phase, indicating a shift from overt revision to more internalized, token-efficient reasoning.}
    % \vspace{-10pt}
    \label{fig:behavior-evolution}
\end{figure}

% \subsection{Evolution of Reasoning Behaviors in \ours{} Training}\label{sec:behavioral_analysis}
\subsection{Evolution of Reasoning Behaviors: \ours{} Internalizes Self-Revision}\label{sec:behavioral_analysis}
% \textbf{\ours{} Internalizes Verbatim Self-Revision into Proactive Reasoning.}
% \paragraph{Behavioral Analysis:} 
%Self-Distillation Internalizes Revision into Compact Reasoning
% The accuracy gains in the previous section raise a natural question: does self-distillation simply improve performance, or does it change \emph{how} the model reasons? We find evidence for the latter.
% We study how the model's generation behavior changes over the course of training in \ours{}.

\looseness-1As shown in \Cref{fig:self-revision}, the \ours{} model after \opsd{} training produces substantially shorter generations, suggesting that it might have internalized some
formerly explicit reasoning. Here, we take a closer look at how this generation behavior evolves over \ours{} training.
Figure~\ref{fig:behavior-evolution} \text{(Right)} tracks two statistics: average response length and frequency of explicit self-revision keywords (e.g., ``Wait'' ``let me start over''; see Appendix \ref{app:keywords} for the full keyword list). During the \sft{} phase, both metrics rise substantially. The model learns to solve problems by actively judging its own response and applying self-correction. During \opsd{} phase, the trend reverses: both response length and self-revision keyword frequency fall steadily, while accuracy continues to improve. The resulting model only uses approximately half as many tokens as the \sft{} model at better accuracy.

\looseness-1We qualitatively illustrate this two-stage pattern via example traces from different models in \Cref{fig:behavior-evolution} (Left). The base model answers with a false claim. The \sft{} model reached the correct answer by backtracking with the exact transition phrase seen in training (``Wait, this is wrong. Let me start over.’’). By contrast, the \ours{} model, without any backtracking, anticipates the same pitfall and directs proactively to the correct answer.
% The qualitative shift is visible in individual traces (Figure~\ref{fig:behavior-evolution} (Left)). 
% While the \sft{} model backtracks with verbatim transition keywords from the training data (``Wait, this is wrong. Let me start over.''), the \ours{} model anticipates the same pitfall and redirects proactively (``coordinates might be a bit tricky. Alternatively, maybe it's easier to think of the grid as a lattice''). 
Note that \opsd{} does not reduce the revision capability acquired in \sft{}. The model still achieves a 5.3\% gain in the Generate-then-Revise evaluation (Figure~\ref{fig:self-revision}), but it has successfully internalized some of that capability into a more directed, pitfall-aware attempt.

\subsection{Ablations on \ours{} Design}
\label{sec:ablations}

We examine several design choices in \ours{}.
% \textbf{Replacing Binary Reward with Self-Verification.} 

\looseness-1\textbf{Loss Terms in \sft{} Objective.} We ablate the two terms in \lsrt{} by training with only \lgen{} or only \lrev{} (\Cref{tab:phase1-ablation}). Neither alone matches full \sft{}: \lgen{} preserves stronger generation but weak self-revision, while \lrev{} improves revision but barely boosts generation. The two terms are therefore complementary: \lrev{} elicits self-revision, and \lgen{} transfers it to stronger generation.
% The SRT objective \lsrt{} (Section~\ref{sec:method}) combines two terms: the first term \lrev{} trains the model to produce a revision or rephrase conditioned on the outcome; the second term \lgen{} trains it to generate better responses by actively evaluating its current response and self-correcting. We ablate this by training with each term alone. If the self-correction term is redundant, removing it should not hurt Phase~2 performance; if the conditioned revision term is redundant, the model should still learn to revise from the self-correction objective alone. \yh{compare with SRT performance and figure 3}

\looseness-1\textbf{\opsd{} Phase without \sft{}.} \ours{} requires the model to have certain self-revision ability, which is unlocked only after the \sft{} phase (\Cref{fig:self-revision}). To further demonstrate its necessity, we apply the \opsd{} phase directly to \qwen{} without prior self-revision training. \Cref{tab:phase2-ablation} shows that this Phase-2-only variant yields only marginal improvements over the base model, and does not improve self-revision ability. Therefore, the \sft{} phase is necessary for \ours{} to be effective.

\looseness-1\textbf{Data Split Between Phases.} We study how to allocate a fixed training data budget between the \sft{} and \opsd{} phases. As shown in \Cref{tab:data-split-ablation}, giving more data to \sft{} slightly improves the \sft{} model, but does not improve the final performance. Instead, the best result is obtained by assigning more data to the \opsd{} phase, suggesting that once \sft{} has unlocked self-revision capability, \opsd{} phase uses additional data more effectively to distill and refine this behavior into stronger generation.

% Allocating more data to SRT yields a stronger reviser but leaves fewer examples for distillation; conversely, too little SRT data may produce a reviser too weak to provide useful token-level supervision. We expect a non-monotonic relationship: at one extreme, insufficient SRT data yields a weak reviser whose noisy supervision limits Phase~2; at the other, over-allocating to SRT starves Phase~2 of fresh examples. We vary the split across \sk{todo: insert} to identify the tradeoff.

%\section{Discussion (Alt version)}

%\clearpage

%The error-aware structure of revision traces---supervising the trajectory from mistake to recovery, not just the correct endpoint---appears to be the key ingredient: SRT's largest gains are on out-of-distribution benchmarks where error recovery is most valuable, and SFT on expert traces actually degrades performance where the distribution gap is largest. Self-distillation then compresses this two-attempt behavior into shorter, single-attempt reasoning while preserving accuracy and revision capability (Figure~\ref{fig:self-revision}). Whether the resulting model genuinely anticipates errors or merely produces shorter outputs remains an open empirical question.

\section{Discussion}
\label{sec:discussion}
% \looseness-1\ours{} shows that a model's own revision capability, once elicited with a small amount of supervised training on self-generated revisions, can transform sparse binary outcome feedback into dense token-level supervision without requiring an external teacher. Across two model families and eight benchmarks, this yields consistent gains of at least \(10\%\) over the base models and outperforms all baselines under the same data budget (\Cref{tab:main_results}); independent experiments on a different task, model, and pipeline variant corroborate both the findings and key design principles (\Cref{app:countdown}).
% More broadly, our results suggest a different view of self-improvement: rather than relying on stronger teacher-generated reasoning responses, a model can learn to convert its own reward-conditioned revisions into token-level supervision.

\looseness-1 \textbf{Conclusion.} \ours{} shows that a model’s revision ability, elicited with a small amount of supervised training on self-generated revisions, can transform sparse binary outcome reward into dense token-level supervision without requiring an external teacher or high-quality demonstrations. 
Across two model families and eight benchmarks, this yields consistent \(10\%+\) gains over the base models and outperforms all baselines under the same training sample budget (\Cref{tab:main_results}). 
We also show experiments on more broader tasks and models in Appendix \ref{app:countdown}. 
Extensive ablations and analyses show that \ours{} also leads to improved revision capability which can lead to further gains on accuracy.
We hope that by eliminating the need for high-quality demonstrations, we enable future work to apply self-distillation to broader domains.

\looseness-1\textbf{Limitations and Open Directions.} Our study focuses on instruct models that generate  short and concise responses, consistent with prior self-distillation works. An important next step is extending self-distillation to \emph{thinking models}, which produce long, exploratory chains of thought. This is challenging because such responses may include false starts and partial corrections that are not mistakes, making it difficult to distinguish productive exploration from genuine errors and assign credit beyond local token decisions. We provide preliminary evidence of this challenge in \Cref{app:opsd-thinking}: applying SDFT to Qwen3-4B with thinking enabled during training hurts performance on competition math benchmarks.

\looseness-1Furthermore, we focus on verifiable domains such as math and coding. Extending \ours{} to domains without verifiable rewards remains an open problem. One promising direction is to define rewards using meta-cognitive signals \citep{didolkar2024metacognitive,didolkar2025metacognitive,shao2025dr, he2025skilltargetedadaptivetraining, he2025skilltargetedadaptivetraining}, such as consistency or self-correction. We leave this to future work.

\bibliography{colm2026_conference}
\bibliographystyle{colm2026_conference}

\newpage
\appendix

%\listofappendices
\section*{Appendix Content}
\vspace{0.5em}
\noindent
\hyperref[sec:related]{A \hspace{0.5em} Related Work} \dotfill \pageref{sec:related} \\[6pt]
\hyperref[app:methoddesign]{B \hspace{0.5em} Method Design} \dotfill \pageref{app:methoddesign} \\[2pt]
\hyperref[app:methoddesign]{\hspace{1.7em} B.1 \hspace{0.5em} \ours{} Training Algorithm} \dotfill \pageref{app:methoddesign} \\[2pt]
\hyperref[app:methoddesign]{\hspace{1.7em} B.2 \hspace{0.5em} Comparison to Existing Methods} \dotfill \pageref{app:methoddesign} \\[6pt]
\hyperref[app:training-data]{C \hspace{0.5em} Experiment Details} \dotfill \pageref{app:training-data} \\[2pt]
\hyperref[app:training-data]{\hspace{1.7em} C.1 \hspace{0.5em} Curating \ours{} training data} \dotfill \pageref{app:training-data} \\[2pt]
\hyperref[app:compare-budget]{\hspace{1.7em} C.2 \hspace{0.5em} Comparing Sampling Budgets} \dotfill \pageref{app:compare-budget} \\[2pt]
\hyperref[app:hyperparameters]{\hspace{1.7em} C.3 \hspace{0.5em} Hyperparameters} \dotfill \pageref{app:hyperparameters} \\[6pt]
\hyperref[app:additionalresults]{D \hspace{0.5em} Additional Results} \dotfill \pageref{app:additionalresults} \\[2pt]
\hyperref[app:additionalresults]{\hspace{1.7em} D.1 \hspace{0.5em} Pass@8 Performance on Math Benchmarks} \dotfill \pageref{app:additionalresults} \\[2pt]
\hyperref[app:revision-token-tradeoff]{\hspace{1.7em} D.2 \hspace{0.5em} SDFT Under Final-Answer-Only Supervision} \dotfill \pageref{app:sdsf-final-answers} \\[2pt]
\hyperref[app:revision-token-tradeoff]{\hspace{1.7em} D.3 \hspace{0.5em} Detailed Self-Revision Statistics} \dotfill \pageref{app:revision-token-tradeoff} \\[2pt]
\hyperref[app:additionalresults]{\hspace{1.7em} D.4 \hspace{0.5em} GRPO Configuration Comparison} \dotfill \pageref{app:additionalresults} \\[6pt]
\hyperref[app:analysisofours]{E \hspace{0.5em} Analysis of \ours{}} \dotfill \pageref{app:analysisofours} \\[2pt]
\hyperref[app:keywords]{\hspace{1.7em} E.1 \hspace{0.5em} Self-Revision Keyword Analysis} \dotfill \pageref{app:keywords} \\[2pt]
\hyperref[app:analysisofours]{\hspace{1.7em} E.2 \hspace{0.5em} Ablation Studies on \ours{} Method Design} \dotfill \pageref{app:analysisofours} \\[6pt]
\hyperref[app:opsd-thinking]{F \hspace{0.5em} SDFT with Thinking Models} \dotfill \pageref{app:opsd-thinking} \\[6pt]
\hyperref[app:countdown]{G \hspace{0.5em} Self-Revision Design Choices} \dotfill \pageref{app:countdown} \\[2pt]
\hyperref[app:countdown-sources]{\hspace{1.7em} G.1 \hspace{0.5em} Effect of Training Data Source} \dotfill \pageref{app:countdown-sources} \\[2pt]
\hyperref[app:countdown-ood]{\hspace{1.7em} G.2 \hspace{0.5em} Self-Revision as Initialization for RL} \dotfill \pageref{app:countdown-ood} \\
\clearpage

\section{Related Work}\label{sec:related}
\paragraph{On-Policy Distillation and Self-Distillation.}
Knowledge distillation trains a student to match a teacher's output distribution either at the token level via soft targets \citep{hinton2015distilling} or at the sequence level \citep{kim2016sequence}. 
However, such off-policy training introduces a train-inference distribution mismatch that leads to compounding errors at test time \citep{bengio2015scheduled, chen2025retaining}.
On-policy distillation (OPD) \citep{agarwal2024onpolicy, gu2024minillm, boizard2024uld, minixhofer2025cross} addresses this by using student rollouts with per-token supervision from the teacher.

A key limitation of OPD is the need for an external model during training to score student traces. 
Recent \emph{self-distillation} methods \citep{furlanello2018born, zhang2019teacher} eliminate this by conditioning the student on privileged information to serve as its own teacher \citep{askell2021general, snell2022learning}, where the privileged 
context can include reasoning traces \citep{zhao2026self}, in-context demonstrations \citep{shenfeld2026selfdistillationenablescontinuallearning}, or conciseness prompts \citep{sang2026onpolicy}.
Still, such works assume access to high-quality signals from external sources that are often costly to obtain, and those that remove the need for an external teacher entirely require rich, dense environment feedback \citep{hubotter2026reinforcement}.
In contrast, \ours{} only requires binary correctness signals of the student's response and generates its own supervision through self-revision.

\paragraph{Self-Training and Self-Refinement.}
Self-training methods improve models by iteratively generating and filtering their own data. 
Prior work bootstraps reasoning by fine-tuning on self-generated rationales that yield correct answers \citep{zelikman2022star, yuan2023scalingrelationshiplearningmathematical, singh2023rest, yuan2024selfrewarding}.
In particular, STaR \citep{zelikman2022star} iteratively generates and filters for correct rationales, discarding incorrect reasoning entirely.
By contrast, SRT retains the incorrect attempt as context and trains on the full mistake-to-correction trajectory.
Consistent with this,
recent evidence suggests that productive reasoning behaviors (e.g., verification, backtracking, etc.) matter more for self-improvement than answer correctness \citep{gandhi2025cognitive}.
Separately, prompting-based self-refinement methods show that LLMs are capable of critiquing and revise their own outputs at inference time \citep{madaan2023selfrefine, shinn2023reflexion}, though such approaches yield limited gains without external feedback and do not update model weights \citep{huang2023selfcorrect, kamoi2024selfcorrection}.
Training-based approaches address this by using RL or supervised learning to internalize correction behavior \citep{kumar2024score, havrilla2024glore}, but still require multi-turn generation at inference time.
\ours{} bridges both avenues by training the model to self-revise and then distilling the revision-informed distribution back into single-pass generation, eliminating multi-turn correction at test time.
%\sk{todo: connection to process reward models (PRMs); probably relevant since token-kl signal in figure 5 is functionally similar to what a PRM would provide}

\paragraph{Connection to Process Reward Models}
The per-token KL signal from the reviser is functionally analogous to process reward models (PRMs) \citep{lightman2024lets, wang-etal-2024-math}, in that both provide localized supervision over intermediate reasoning rather than only the final outcome.
PRMs train a separate model to assign step-level scores to individual reasoning steps.
In \ours{}, the token-level divergence $D^{(t)}_{KL}$ similarly concentrates on a sparse subset of tokens associated with errors in the response (Figure \ref{fig:credit-assignment}), so that the log-probability gap at each token both penalizes erroneous reasoning and redirects probability mass towards plausible alternatives. The main difference is cost and supervision: PRMs typically require either step-level correctness annotations \citep{lightman2024lets} or rollout-based estimates from search or sampling \citep{wang-etal-2024-math}, together with a separately trained reward model, whereas our reviser's signal arises by conditioning the same model on its own response and a binary outcome, without step-level annotation or an auxiliary reward model. Whether this implicit token-level signal can complement or serve as a cheaper alternative to trained PRMs in settings such as guided tree search remains an interesting direction for future work. 
%\sk{todo: will add more prm refs}

% \sk{todo: also cite llm behavior noah goodman paper; also howards paper for on policy data}\edward{which papers are these?}

% \sk{https://arxiv.org/abs/2503.01307, https://arxiv.org/abs/2510.18874}
% \sk{i think we can also include some small part about self-distillation ina "traditional" sense eg born again networks? https://arxiv.org/abs/1805.04770}

% \sk{sdpo also includes the binary reward (+env feedback) in context, so we can include a short comparison to that here and say that what makes our work different: (1) we teach the model to revise in the first place (2) sdpo doesnt include the unsuccessful attempt in the context (instead it include the correct student rollout in context for the group)}

\clearpage
\section{Method Design}
\label{app:methoddesign}
% \yh{show that rephrase token decreases}
%i swapped the order of these subsections to make it fit proprely (but feel free to revert if oyu think the other ordering makes more sense)

\subsection{\ours{} Training Algorithm}
% \yh{modify notations}
\Cref{alg:pipeline} summarizes the full two-phase training procedure of \ours{} (\Cref{sec:method}). In Phase 1 (\sft{}), the model first learns outcome-conditioned self-revision: for each sampled response, we use the binary verifier to determine whether the response is correct, construct a control prompt that either triggers rephrasing (r=1) or restarting (r=0), and keep only revised responses that are verified as correct to form the self-revision dataset \(\mathcal{D}_{\textsc{revision}}\). The model is then trained on this dataset with the \sft{} objective, which jointly improves its ability to revise and to generate stronger responses. In Phase 2 (\opsd{}), we initialize the student from the \sft{} model and perform on-policy self-distillation: the current student generates a response, the fixed \sft{} reviser conditions on that response and its binary outcome to define a teacher distribution, and the student is updated to match this distribution via KL minimization. In this way, \ours{} first elicits explicit self-revision behaviors and then distills them back into more compact, token-efficient generation.

\begin{algorithm}[htbp]
\caption{\ours{} Training Pipeline}
\label{alg:pipeline}
\begin{algorithmic}[1]
\small
\Require Base model $\pi_\theta$, dataset $\mathcal{D}=\{(x,a)\}_{i=1}^N$, binary verifier $r(y,a)\in\{0,1\}$
\Statex \vspace{-0.5em}

\Statex \textbf{$\vartriangleright$ Phase 1 (\sft{}): Self-Revision Training}
\For{each $(x,a)\in\mathcal{D}$}
    \State Sample initial response $y_{\text{init}} \sim \pi_\theta(\cdot \mid x)$
    \State Compute reward $r \gets r(y_{\text{init}}, a)$
    \State Construct control prompt
    \[
    P_r \gets
    \begin{cases}
    \text{``Let me rephrase the above solution.''}, & r=1,\\
    \text{``Wait, this response is not correct, let me start over.''}, & r=0
    \end{cases}
    \]
    \State Generate revised response $y_{\text{revised}} \sim \pi_\theta(\cdot \mid x, y_{\text{init}}, P_r)$
    \State Add $(x, y_{\text{init}}, P_r, y_{\text{revised}})$ to $\mathcal{D}_{\textsc{revision}}$ \textbf{if} $r(y_{\text{revised}}, a)=1$
\EndFor
\State \textcolor{darkbluegreen}{\textbf{Train $\pi_\theta$ on $\mathcal{D}_{\textsc{revision}}$ with $\mathcal{L}_{\text{\sft{}}}$ to obtain $\pi_{\theta_{\sft{}}}$}}

\Statex
\Statex \textbf{$\vartriangleright$ Phase 2 (\opsd{}): On-Policy Self-Distillation via Revision Feedback}
\State Initialize $\pi_\theta \gets \pi_{\theta_{\sft{}}}$
\Repeat
    \State Sample $(x,a)\sim \mathcal{D}$
    \State Sample student response $y_{\text{init}} \sim \pi_\theta(\cdot \mid x)$
    \State Compute reward $r \gets r(y_{\text{init}}, a)$
    \State Construct control prompt $P_r$ as in Phase 1
    \State \textcolor{darkbluegreen}{\textbf{Define student policy $\pi_S(\cdot) \gets \pi_\theta(\cdot \mid x)$}}
    \State \textcolor{darkbluegreen}{\textbf{Define teacher policy $\pi_T(\cdot) \gets \pi_{\theta_{\sft{}}}(\cdot \mid x, y_{\text{init}}, P_r)$}}
    \State Update $\theta$ by minimizing
    \[
    \mathcal{L}_{\text{\opsd{}}}(\theta)
    = \KL\!\left(\pi_S \,\|\, \stopgrad(\pi_T)\right)
    \]
\Until{convergence}
\end{algorithmic}
\end{algorithm}

\clearpage 

\subsection{Comparison to Existing Methods}
\begin{table}[h!]
\centering
\begin{tabular}{lcccc}
\toprule
\textbf{Method} & \textbf{Sampling} & \textbf{Signal} & \textbf{Teacher} & \makecell{\textbf{Teacher can condition} \\ \textbf{on wrong attempt}} \\
\midrule
SFT / Distillation & \xmark~off-policy & \cmark~dense & \xmark~external & --- \\
On-Policy Distillation & \cmark~on-policy & \cmark~dense & \xmark~external & --- \\
RLVR (e.g., GRPO) & \cmark~on-policy & \xmark~sparse & --- & --- \\
OPSD / SDFT / SDPO & \cmark~on-policy & \cmark~dense & \cmark~self & \xmark \\
\midrule
\textbf{\ours{} (ours)} & \cmark~on-policy & \cmark~dense & \cmark~self & \cmark \\
\bottomrule
\end{tabular}
\caption{Comparison of post-training methods for LLMs. \ours{} is the only self-distillation method whose teacher can condition on incorrect student attempts, enabling it to convert sparse binary rewards into dense token-level supervision without requiring gold demonstrations. 
% \sk{feel free to comment out this table if it doesn't add value}
}
\label{tab:comparison}
\end{table}
\Cref{tab:comparison} situates \ours{} among existing post-training methods. Standard SFT and distillation rely on dense supervision but are off-policy and require external demonstrations, while RLVR methods such as GRPO are on-policy but optimize only sparse outcome rewards. Recent self-distillation methods such as OPSD, SDFT, and SDPO combine on-policy sampling with dense self-supervision, but their teacher does not explicitly condition on the student’s failed trajectory. In contrast, \ours{} is the only method in this comparison that is simultaneously on-policy, self-distilled, and able to condition the teacher on an incorrect attempt, allowing it to transform binary outcome rewards into targeted token-level supervision without requiring gold reasoning traces.

\clearpage
\section{Experiment Details}
\subsection{Curating \ours{} training data} \label{app:training-data}
In \ours{}, we trained the models on 15K training data from \openmath{} and 15K data from Codeforces separately, with 6K in Phase 1 (\sft{}) and 9K in Phase 2 (\opsd{}).

In \sft{} phase, we curate 6K self-revision training data from 10K question-answer pairs in \openmath{} (or \cf{}), through the following pipeline:
\begin{enumerate}
    \item \textbf{Initial sampling:} Select the first 10K questions in \openmath{} (or \cf{}), and sample $1$ initial model responses $y_{\text{initial}}$ per question $x$,
    \item \textbf{Verification:} Verify the binary reward $r\in\{0,1\}$ for each $y_{\text{initial}}$, and build self-revision prompt $P_r$. The 10K initial responses are roughly split into 5K correct and 5K incorrect responses.
    \item \textbf{Self-Revision:} For each correct initial response, prompt the model to generate $3$ rephrased responses $y_\text{revised}$; For each incorrect initial response, prompt the model to generate $3$ corrected responses $y_\text{revised}$,
    \item \textbf{Filtering:} Keep traces $(x,y_{\text{initial}},P_r,y_\text{revised})$ where $y_\text{revised}$ reaches a correct final answer. The resulting training data contain 6K self-revision traces.
\end{enumerate}

In \opsd{} phase, we directly sample an additional 9K question-answer pairs from \openmath{} (or \cf{}) as training data.

\subsection{Comparing Sampling Budgets} \label{app:compare-budget}

One feature of \ours{} is sample efficiency. In \Cref{tab:budget-comparison}, we compare the sampling budget of \ours{} with baseline on-policy training methods, including rejection fine-tuning (RFT), GRPO, and SDFT. We calculate the sampling budget of each methods as follows:\\

\begin{itemize}
    \item \textbf{RFT:} total \# generations = 15K (questions) $\times$ 4 (response attempts per question) = 60K.
    \item \textbf{GRPO:} total \# generations = 15K $\times$ 4 (rollout/question) = 60K.
    \item \textbf{SDFT:} total \# generations = 15K $\times$ 4 (rollout/question) = 60K.
    \item \textbf{\sft{} phase:} total \# generations = 10K (questions) $\times$ 1 (initial response attempts per question) + 5K (correct initial responses) $\times$ 3 (rephasing attempts per correct initial response) + 5K (incorrect initial responses) $\times$ 3 (self-correction attempts per incorrect initial response) = 40K. 
    \item \textbf{\opsd{} phase:} total \# generations = 9K $\times$ 1 (rollout/question) = 9K. 
    \item \textbf{\ours{}:} 40K (\sft{} phase) + 9K (\opsd{} phase) = 49K.
\end{itemize}

We trained the GRPO baseline for 1 epoch on 15K training qesutions, with 4 generations per question \footnote{We also compare to the 8-rollout GRPO variants in \Cref{tab:grpo_ablation}.}.
\begin{table}[h]
\centering
\small
\begin{tabular}{lcccc}
\toprule
\multirow{2}{*}{Method} & \# Training & \# Training rollouts & \# Training & \textbf{Total} \\
 & questions & per question & epochs & \textbf{\# generations} \\
\midrule
% SFT & 15K & 2 & 0 \\
RFT & 15K & - & 1 & \textbf{60K} \\
GRPO & 15K & 4 & 1 & \textbf{60K} \\
SDFT & 15K & 4 & 1 & \textbf{60K} \\
\midrule
\sft{} phase & 6K & - & 1 & \textbf{40K} \\
\opsd{} phase & 9K & 1 & 1 & \textbf{9K} \\
% \midrule
\textbf{\ours{}} & 15K & - & 1 & \textbf{49K} \\
\bottomrule
\end{tabular}
\caption{Comparison of training data and generation budgets across different training paradigms. For RFT, GRPO, and SDFT, we assume 15K training questions and 4 sampled rollouts per question, resulting in 60K total generations before any filtering. In our method, the \sft{} phase starts from 10K seed questions, generates 1 initial response and 3 self-revisions per question (40K generations in total), and retains 6K correct self-revision traces for training. The \opsd{} phase directly uses 9K additional question-answer pairs, requiring 9K generations. In total, \ours{} uses 15K training questions and 49K generations.}
\label{tab:budget-comparison}
\end{table}

% \sk{maybe i missed this, but can we also add some details about which 6k questions are used for revisions, and what the sampling budget is like (ie how many first attempts per question is the model allowed to make? and how many revisions + rephrasings can the model make per first attempt? and how does compare to the sampling budget for grpo/dapo? from my understanding, we claim that osrd (phase 2) is more sample efficient than rlvr methods, but i think we should confirm that both phase 1 + phase 2 together can be more sample efficient than rlvr}

We also estimate the total token budget. At approximately 3.7K tokens per response, the sampling budgets for RFT and GRPO are each roughly 222M tokens. For \ours{}, the SRT data collection phase generates approximately 148M tokens (10K initial responses plus 30K revisions), and the Self-Distillation phase generates at most 76.5M tokens (9K rollouts at up to 8.5K tokens each, though rollout length decreases during training). The total sampling budget for \ours{} is thus at most 225M tokens, comparable to baselines despite achieving substantially stronger performance.

\paragraph{Axis 1: Sampling budget (total completion tokens generated).}
\begin{itemize}
    \item \textbf{RFT:} $60\text{K} \times 3.7\text{K} \approx 222\text{M}$ tokens
    \item \textbf{GRPO:} $60\text{K} \times 3.7\text{K} \approx 222\text{M}$ tokens
    \item \textbf{\ours{} Phase 1 (SRT data collection):}
    \begin{itemize}
        \item 10K initial responses $\times\; 3.7\text{K} = 37\text{M}$
        \item 30K revisions $\times\; 3.7\text{K} = 111\text{M}$
        \item Subtotal: $148\text{M}$ tokens
    \end{itemize}
    \item \textbf{\ours{} Phase 2 (Self-Distillation):}
    \begin{itemize}
        \item 9K rollouts $\times\; 8.5\text{K} \leq 76.5\text{M}$ tokens (upper bound; rollout length decreases during this phase per Figure~4)
    \end{itemize}
    \item \textbf{\ours{} total:} $\leq 224.5\text{M}$ tokens
\end{itemize}

\paragraph{Axis 2: Training budget (completion tokens in forward pass for loss computation).}
\begin{itemize}
    \item \textbf{GRPO:} $60\text{K} \times 3.7\text{K} \approx 222\text{M}$ tokens (probably more than this since avg response length increases to 4.4K by the end of GRPO)
    \item \textbf{RFT:} $\leq$ 222M (same sampling budget as GRPO but with incorrect samples dropped)
    \item \textbf{\ours{} SRT training} (6K traces, two losses per trace):
    \begin{itemize}
        \item $\mathcal{L}_{\text{revision}}$: $6\text{K} \times 3.7\text{K} = 22\text{M}$
        \item $\mathcal{L}_{\text{generation}}$: $6\text{K} \times 7.4\text{K} = 44\text{M}$ (uses $2 \times 3.7K$ since $y' = [y_{\text{init}}, P_r, y_{\text{revised}}]$)
        \item Subtotal: $66\text{M}$ tokens
    \end{itemize}
    \item \textbf{\ours{} Self-Distillation} (student + teacher forward pass per step):
    \begin{itemize}
        \item Student: $9\text{K} \times 8.5\text{K} = 76.5\text{M}$
        \item Teacher: $9\text{K} \times 8.5\text{K} = 76.5\text{M}$
        \item Subtotal: $\leq 153\text{M}$ tokens (probably less than this since avg response length goes down during self-distillation phase)
    \end{itemize}
    \item \textbf{\ours{} total:} $\leq 219\text{M}$ tokens
\end{itemize}

\subsection{Hyperparameters} \label{app:hyperparameters}
We attach the hyperparameters used for \qwen{} in SFT, RFT, and \ours{} \sft{} phase in \Cref{tab:sft_hparams}, those used in GRPO in \Cref{tab:grpo_hparams}, and those used in SDFT and \ours{} \opsd{} phase in \Cref{tab:distillation_hparams}.
\begin{table}[h!]
\centering
\small
\setlength{\tabcolsep}{10pt}
\renewcommand{\arraystretch}{1.15}
\fontsize{8.5pt}{8.5pt}\selectfont
\begin{tabular}{p{5.5cm}p{7cm}}
\toprule
\textbf{Parameters} & \textbf{SFT} \\
\midrule

\textbf{General} & \\
Model & Qwen/Qwen3-4B-Instruct-2507 \\
Thinking & False \\
% WandB mode & offline \\

\midrule
\textbf{Data} & \\
% Training data & \openmath{} \\
% Training file & \texttt{code\_Qwen3-4B\_15k\_samples.json} \\
% Sample size & 6000 \\
Prompt format & Chat template (system + user) \\
Completion-only loss & True \\
Max. sequence length & 32768 \\
Max. response length & 32768 \\

\midrule
\textbf{Batching} & \\
Per-device train batch size & 1 \\
Per-device eval batch size & 1 \\
Gradient accumulation steps & 1 \\
Effective global batch size & 4 \\
% Number of GPUs & 4 \\

\midrule
\textbf{Optimization / Training} & \\
Optimizer & AdamW \\
Learning rate & $5 \times 10^{-6}$ \\
Scheduler & Cosine \\
Warmup ratio & 0.05 \\
Num. epochs & 3 \\
Weight decay & $1 \times 10^{-4}$ \\
Adam $\beta_1$ & 0.9 \\
Adam $\beta_2$ & 0.95 \\
Precision & bfloat16 \\
Gradient checkpointing & True \\
Gradient accumulation sync each batch & True \\
Use Liger kernel & True \\

% \midrule
% \textbf{Evaluation / Checkpointing} & \\
% Evaluation strategy & no \\
% Logging steps & 1 \\
% Save steps & 1000 \\
% Push to hub & False \\

\midrule
\textbf{Parallelism / FSDP} & \\
% Launcher & \texttt{torchrun} \\
FSDP mode & \texttt{full\_shard auto\_wrap} \\
FSDP wrapped layer & \texttt{Qwen3DecoderLayer} \\

\bottomrule
\end{tabular}
\caption{Hyperparameters used for SFT/ RFT baselines and \ours{} \sft{} phase on \qwen{}.}
\label{tab:sft_hparams}
\end{table}
\begin{table}[h!]
\centering
\small
\setlength{\tabcolsep}{10pt}
\renewcommand{\arraystretch}{1.15}
\fontsize{8pt}{8.5pt}\selectfont
\begin{tabular}{p{5.5cm}p{7cm}}
\toprule
\textbf{Parameters} & \textbf{GRPO} \\
\midrule

\textbf{General} & \\
Model &  Qwen/Qwen3-4B-Instruct-2507\\
Thinking & False \\

\midrule
\textbf{Data} & \\
% Training data & OpenR1-Math\\
% Prompt key & \texttt{prompt} \\
% Reward function key & \texttt{data\_source} \\
Max. prompt length & 1024 \\
Max. response length & 8192 \\
Train batch size & 64 \\
Validation batch size & -- \\
Shuffle & True \\
Truncation & left \\

\midrule
\textbf{Batching} & \\
Rollouts per prompt ($n$) & 8 \\
PPO mini-batch size & 64 \\
PPO micro-batch size per GPU & 1 \\
Logprob micro-batch size per GPU & 1 \\
Max. sequences per rollout batch & 64 \\
Max. batched tokens & 16384 \\

\midrule
\textbf{Rollout / Generation} & \\
Inference engine & vLLM \\
Rollout mode & sync \\
Temperature & 0.7 \\
Top-$p$ & 1.0 \\
Top-$k$ & -- \\
Sampling & True \\
% Validation temperature & 0 \\
% Validation samples per prompt & 1 \\
% Validation sampling & False \\
Max new tokens & 8192 \\
Tensor parallel size & 2 \\
% GPU memory utilization & 0.4 \\
Enable chunked prefill & True \\
% Load format & dummy \\

\midrule
\textbf{Actor / Policy Optimization} & \\
Strategy & FSDP \\
PPO epochs & 1 \\
Clip ratio & 0.2 \\
Entropy coefficient & 0.001 \\
Use KL loss & True \\
KL loss coefficient & 0.001 \\
KL loss type & low-var KL \\
Gradient clip norm & 1.0 \\
Dynamic batch size & False \\
% Max token length per GPU & 12500 \\

\midrule
\textbf{Reference Model} & \\
Strategy & FSDP \\
% Logprob micro-batch size per GPU & 1 \\
Dynamic batch size & False \\
Sequence parallel size & 1 \\

% \midrule
% \textbf{Critic / Reward Model} & \\
% Critic enabled & False \\
% Reward model enabled & False \\

\midrule
\textbf{Parallelism} & \\
Number of GPUs & 8 \\
Number of nodes & 1 \\
Sequence parallel & 1 \\
Actor strategy & FSDP \\
Reference strategy & FSDP \\
Gradient checkpointing & True \\

\midrule
\textbf{Training} & \\
Optimizer & AdamW \\
Learning rate & $1 \times 10^{-6}$ \\
Scheduler & Constant \\
Warmup steps ratio & 0.0 \\
Weight decay & 0.01 \\
Precision & bfloat16 \\
NCCL timeout & 1800 \\

\bottomrule
\end{tabular}
\caption{Hyperparameters used in GRPO baseline on \qwen{}.}
\label{tab:grpo_hparams}
\end{table}

\begin{table}[h!]
\centering
\small
\setlength{\tabcolsep}{8pt}
\renewcommand{\arraystretch}{1.15}
\fontsize{8pt}{8.5pt}\selectfont
\begin{tabular}{p{3.5cm}p{4cm}p{4cm}}
\toprule
\textbf{Parameters} & \textbf{\ours{}} & \textbf{SDFT} \\
\midrule

\textbf{General} & & \\
Policy Model & Qwen/Qwen3-4B-Instruct-2507 & Qwen/Qwen3-4B-Instruct-2507 \\
Teacher model & Qwen/Qwen3-4B-Instruct-2507 & Qwen/Qwen3-4B-Instruct-2507 \\
Thinking & False & False \\
% Seed & 42 & 42 \\

\midrule
\textbf{Data} & & \\
% Training data & OpenR1-Math & OpenR1-Math \\
Max. teacher prompt length & 32768 & 32768 \\
Max. response length & 8192 & 8192 \\

\midrule
\textbf{Batching} & & \\
Prompts per step & 128 & 128 \\
Generations per prompt & 1 & 1 \\
Train global batch size & 128 & 128 \\
Train micro batch size & 1 & 1 \\
Generation batch size & 64 & 128 \\
Logprob batch size & 1 & -- \\
Max rollout turns & 1 & 1 \\

\midrule
\textbf{Rollout / Generation} & & \\
Inference engine & vLLM & vLLM \\
Temperature & 1.0 & 1.0 \\
Top-$p$ & 1.0 & 1.0 \\
Top-$k$ & -- & -- \\
Max new tokens & 8192 & 8192 \\
vLLM tensor parallel size & 4 & 1 \\

\midrule
\textbf{Validation} & & \\
Validation batch size & 64 & -- \\
Validation period & 20 & -- \\
% Validate at start & True & -- \\
% Validate at end & True & -- \\

\midrule
\textbf{Distillation loss} & & \\
% KL type & mixed & -- \\
% Mixed KL weight & 0.5 & -- \\
Top-$K$ distillation & 64 & -- \\
Zero outside top-$K$ & False & -- \\

\midrule
\textbf{Parallelism} & & \\
Number of GPUs & 4 & 4 \\
Policy tensor parallel size & 4 & -- \\
Policy context parallel size & 1 & -- \\
Teacher tensor parallel size & 4 & -- \\
Teacher context parallel size & 1 & -- \\
Sequence parallel & False & False \\
Activation checkpointing & True & True \\

\midrule
\textbf{Training} & & \\
Optimizer & AdamW & AdamW \\
Learning rate & $5 \times 10^{-6}$ & $5 \times 10^{-6}$ \\
Scheduler & Linear warmup + constant & Cosine (warmup ratio $=0.1$) \\
Warmup steps & 20 & 20 \\
Weight decay & 0.01 & -- \\
Gradient clip norm & 1.0 & 1.0 \\
Precision & bfloat16 & bfloat16 \\
% Max num epochs & 5 & 3 \\
% Max num steps & 1000 & 200 \\

\bottomrule
\end{tabular}
\caption{Hyperparameters used for \ours{} \opsd{} phase and SDFT on \qwen{}.}
\label{tab:distillation_hparams}
\end{table}

\clearpage
\section{Additional Results}
\label{app:additionalresults}
% \sk{in addition to Figure~\ref{fig:self-revision}, maybe we can include a small table with exact numbers for mean + median tokens per response for instruct model, grpo, opsd, srt, and self-distillation (both for first attempt and revised attempt). We can also include a revision success rate by method table, i.e.\ what fraction of incorrect --> correct flips does each method achieve? If SRT corrects 40\% of wrong first-attempts but grpo corrects only 15\%, that quantifies the ``unlocked revision capability'' claim precisely.}
\subsection{Pass@8 Performance on Math Benchmarks}
\Cref{tab:pass8_results} complements the main results in \Cref{tab:main_results} by showing that the advantages of \sft{} and \ours{} persist under Pass@8 evaluation. The trend is consistent with the avg@16 results: across both Qwen3-4B-Instruct and Olmo-3-7B-Instruct, \sft{} already matches or outperforms strong baselines, and \ours{} further achieves the best overall average performance. These gains on Pass@8 suggest that our method is not merely sharpening the output distribution around a narrow set of solutions, but genuinely making the model’s exploration more well-directed, so that correct reasoning paths are more likely to be discovered across multiple attempts. Overall, the Pass@8 results further reinforce the conclusion that explicit self-revision training improves reasoning quality, and that the second-stage \ours{} training strengthens this effect.

\def\changesize#1{\fontsize{#1}{10.5pt}\selectfont}
\begin{table*}[h]
\centering
\small
\fontsize{9pt}{10.5pt}\selectfont
    \renewcommand{\arraystretch}{1.2}
    \setlength{\tabcolsep}{2.6pt}
\begin{tabular}{lccccccc}
\toprule
 % & \multicolumn{6}{c}{\textbf{Math}} & \multirow{2}{*}{\textbf{Avg.}} \\
\cmidrule(lr){2-7}
 & {AIME24} & {AIME25} & {HMMT25} & {AMOBench} & {OpenR1} & {MATH} & \textbf{Average}\\
\midrule
\textbf{\qwen{}}           & 76.7          & 76.7          & 50.0          & 30.0          & 69.2          & 97.6          & 66.7          \\
SFT                        & 83.3          & 73.3          & 46.7          & 18.0          & 69.2          & 97.0          & 64.6          \\
RFT                        & \textbf{86.7} & \textbf{80.0} & 53.3          & 30.0          & 70.2          & \textbf{98.4} & 69.8          \\
GRPO                       & 80.0          & 70.0          & 46.7          & 28.0          & 70.0          & 98.0          & 65.4          \\
SDFT                       & 80.0          & 70.0          & 50.0          & 26.0          & 70.0          & 97.0          & 65.5          \\
\textbf{\sft{}} (Ours)     & \textbf{86.7} & \textbf{80.0} & 60.0          & 28.0          & 70.2          & 97.6          & 70.4          \\
\textbf{\ours{}} (Ours)    & \textbf{86.7} & \textbf{80.0} & \textbf{63.3} & \textbf{36.0} & \textbf{72.0} & 97.0          & \textbf{72.5} \\
\midrule

\textbf{\olmo{}}           & 80.0          & 73.3          & 46.7          & 16.0          & 60.0          & \textbf{98.0} & 62.3          \\
SFT                        & 83.3          & 66.7          & 46.7          & 18.0          & 55.0          & 96.0          & 60.9          \\
RFT                        & 80.0          & 73.3          & 60.0          & 22.0          & 62.0          & 97.0          & 65.7          \\
GRPO                       & 80.0          & 66.7          & 46.7          & 18.0          & 70.0          & 97.0          & 63.1          \\
SDFT                       & 80.0          & 73.3          & 56.7          & 20.0          & 58.0          & 97.0          & 64.2          \\
\textbf{\sft{}} (Ours)     & \textbf{83.3} & 76.7          & \textbf{66.7} & 20.0          & 66.0          & \textbf{98.0} & 68.4          \\
\textbf{\ours{}} (Ours)    & \textbf{83.3} & \textbf{80.0} & \textbf{66.7} & \textbf{30.0} & \textbf{68.0} & 97.0          & \textbf{70.8} \\
\bottomrule
\end{tabular}
\caption{\textbf{Pass@8 performance comparison of \ours{} and \sft{} (Self-Revision Training, Phase~1) against baseline post-training methods} on math reasoning benchmarks. Across both \qwen{} and \olmo{}, \sft{} and especially \ours{} achieve the strongest overall results.}
% \vspace{-10pt}
\label{tab:pass8_results}
\end{table*}

\subsection{SDFT Under Final-Answer-Only Supervision}
\label{app:sdsf-final-answers}
In \Cref{tab:sdft_final_answers}, we show that SDFT performs much worse when given access only to ground-truth final answers rather than gold solution traces. Its performance remains close to the base model and falls far behind \ours{} across all math benchmarks. These results highlight that existing self-distillation methods do not naturally benefit from final-answer-only supervision, whereas \ours{} is specifically designed to convert such sparse outcome feedback into effective token-level learning signals.
\def\changesize#1{\fontsize{#1}{10.5pt}\selectfont}
\begin{table*}[h]
\centering
\small
\fontsize{9pt}{10.5pt}\selectfont
    \renewcommand{\arraystretch}{1.2}
    \setlength{\tabcolsep}{2pt}
\begin{tabular}{lccccccc}
\toprule
 % & \multicolumn{6}{c}{\textbf{Math}} & \multirow{2}{*}{\textbf{Avg.}} \\
\cmidrule(lr){2-7}
 & {AIME24} & {AIME25} & {HMMT25} & {AMOBench} & {OpenR1} & {MATH} & \textbf{Avg.}\\
\midrule

Base model                                           & 59.6          & 45.8          & 26.7          & 9.8           & 55.8 & 91.0          & 48.1          \\
SDFT                                                 & 63.3          & 47.9          & 29.9          & 9.0           & 57.0 & 91.1          & 49.7          \\
SDFT (final answers only)                           & 62.5          & 47.1          & 29.9          & 9.3           & 57.0 & 91.0          & 49.5          \\
\ours{} (Ours)                                      & \textbf{68.3} & \textbf{60.0} & \textbf{45.4} & \textbf{16.0} & \textbf{60.4} & \textbf{93.6} & \textbf{57.3} \\

\bottomrule
\end{tabular}
\caption{\textbf{SDFT with final-answer-only supervision} on \qwen{} evaluated on math benchmarks. When given access only to final answers rather than gold solutions, SDFT performs only marginally better than the base model and remains far below \ours{}, showing that \ours{} is much more effective under outcome-only supervision.}
\label{tab:sdft_final_answers}
\end{table*}

\subsection{Detailed Self-Revision Statistics}
\label{app:revision-token-tradeoff}

\Cref{tab:revision-token-tradeoff} reports detailed statistics for the Generate-then-Revise evaluation on 1K AIME24 questions with \qwen{} as the base model. For each method, we show the average token cost of the first attempt and the revision, together with the corresponding accuracies before and after revision. We also report the net wrong-to-correct rate, computed as $(\text{Revised Attempt Accuracy} - \text{First Attempt Accuracy}) / (100 - \text{First Attempt Accuracy})$, which measures the fraction of initially incorrect answers that are corrected after revision. These statistics complement \Cref{fig:self-revision} by quantifying both the accuracy gains and the token efficiency of revision across methods.
\vspace{-1em}
\begin{table}[H]
\centering
\setlength{\tabcolsep}{3pt}
\small
\begin{tabular}{lccccc}
\toprule
\multirow{2}{*}{\textbf{Method}} & \textbf{First Attempt} & \textbf{Revised Attempt} & \textbf{First Attempt } & \textbf{Revised Attempt} & \textbf{Correction} \\
 & \textbf{Tokens} & \textbf{Tokens} & \textbf{Accuracy (\%)} & \textbf{Accuracy (\%)} & \textbf{Rate (\%)} \\
\midrule
Base Model & 3708 & 5098 & 59.6 & 60.7 & 2.7 \\
GRPO & 4432 & 5499 & 65.2 & 66.9 & 4.9 \\
SDFT & 3630 & 5099 & 63.3 & 64.7 & 3.8 \\
\sft{}  & 8458 & 8137 & 66.7 & 71.7 & \textbf{15.0} \\
\ours{} & 3518 & 3314 & 68.3 & 73.6 & \textbf{16.7} \\
\bottomrule
\end{tabular}
\caption{Comparison of self-revision effectiveness and token costs across different models on AIME24 with \qwen{} as the base model. The Correction Rate is computed as $(\text{Revised Attempt Accuracy} - \text{First Attempt Accuracy}) / (100 - \text{First Attempt Accuracy})$, i.e., the net fraction of initially incorrect answers corrected after revision.}
\label{tab:revision-token-tradeoff}
\end{table}

\subsection{GRPO Configuration Comparison}
The GRPO baseline in \Cref{tab:main_results} is run under 4 generations per question under a sampling budget matched with \ours{}. Here, we vary the number of generations per question and the number of training epochs in GRPO for \qwen{} on the math domain.

As shown in \Cref{tab:grpo_ablation}, simply increasing the rollout budget in GRPO does not close the gap to our method. Moving from 4 to 8 generations per question (with 0.5 training epochs to match rollout budget) yields only marginal improvements, and in some settings even slightly hurts performance. Extending training to a full epoch with 8 generations (2$\times$ sampling budget as \ours{}) provides some recovery, but the gains remain modest overall.

\def\changesize#1{\fontsize{#1}{10.5pt}\selectfont}
\begin{table*}[h]
\centering
\small
\fontsize{9pt}{10.5pt}\selectfont
    \renewcommand{\arraystretch}{1.2}
    \setlength{\tabcolsep}{2pt}
\begin{tabular}{lccccccc}
\toprule
 % & \multicolumn{6}{c}{\textbf{Math}} & \multirow{2}{*}{\textbf{Avg.}} \\
\cmidrule(lr){2-7}
 & {AIME24} & {AIME25} & {HMMT25} & {AMOBench} & {OpenR1} & {MATH} & \textbf{Avg.}\\
\midrule
GRPO (4 generations, 1 epoch)                       & 62.5 & 50.0 & 30.4 & 11.0 & 62.9 & 93.5 & 51.7 \\
GRPO (8 generations, 0.5 epoch)                    & 58.8 & 50.0 & 30.4 & 9.8  & 61.5 & 93.0 & 50.6 \\
GRPO (8 generations, 1 epoch)                      & 61.7 & 52.1 & 31.3 & 12.0 & \textbf{63.4} & 93.1 & 52.3 \\
SD-Zero (Ours)                                     & \textbf{68.3} & \textbf{60.0} & \textbf{45.4} & \textbf{16.0} & 60.4 & \textbf{93.6} & \textbf{57.3} \\

\bottomrule
\end{tabular}
\caption{\textbf{Comparison with GRPO under larger rollout budgets} on \qwen{} evaluated on math benchmarks. We compare SD-Zero against GRPO variants that use more sampled generations per prompt and different effective training budgets. Even when GRPO is given substantially more exploration through 8 generations, its gains remain limited, whereas SD-Zero achieves the best average performance and outperforms all GRPO settings on most benchmarks. This suggests that the advantage of our method comes not simply from sampling more trajectories, but from turning binary outcome feedback into more informative self-revision-based supervision.}
\label{tab:grpo_ablation}
\end{table*}

\clearpage
\section{Analysis of \ours{}}
\label{app:analysisofours}
\subsection{Self-Revision Keyword Analysis} \label{app:keywords}
Below, we list the self-revision keywords used in our behavioral analysis. These keywords are used in the study described in \Cref{fig:behavior-evolution} to quantify the frequency of explicit self-correction language during training, complementing the analysis of response length and downstream performance. We construct the list by collecting common phrases that models frequently use when revising an earlier reasoning step, such as signaling doubt, detecting an error, restarting, rechecking, or correcting a previous claim. Concretely, the list includes expressions associated with hesitation (e.g., “wait,” “hold on”), error acknowledgment (e.g., “my mistake,” “this is wrong”), and explicit restart or verification behavior (e.g., “let me recheck,” “let me try again,” “let me start over”). While this keyword set is not intended to exhaustively capture all forms of self-correction, it provides a simple and interpretable proxy for tracking overt self-revision behavior across training stages.

\definecolor{boxbg}{HTML}{F7F9FC}
\definecolor{boxframe}{HTML}{4F81BD}
\definecolor{codegreen}{HTML}{1F6F43}

\lstdefinestyle{pythonstyle}{
    language=Python,
    basicstyle=\ttfamily\small,
    keywordstyle=\color{blue}\bfseries,
    stringstyle=\color{darkbluegreen},
    showstringspaces=false,
    breaklines=true,
    frame=none
}
\begin{tcolorbox}[
    colback=lightblue!30,
    colframe=vividblue,
    coltitle=black,
    title=List of Self-Revision Keywords Used in Behavioral Analysis (\Cref{sec:behavioral_analysis}),
    % fonttitle=\bfseries,
    boxrule=0.8pt,
      arc=2pt,
      left=6pt,
      right=6pt,
      top=4pt,
      bottom=4pt
]
\begin{lstlisting}[style=pythonstyle]
REVISION_KEYWORDS = [
    "wait",
    "hold on",
    "actually",
    "on second thought",
    "let me recheck",
    "let's recheck",
    "let me check",
    "let's check",
    "let me recalculate",
    "let's recalculate",
    "let me correct",
    "my mistake",
    "i made a mistake",
    "this is wrong",
    "that is wrong",
    "that's wrong",
    "incorrect",
    "re-evaluate",
    "let's re-evaluate",
    "let me rethink",
    "let's rethink",
    "let me double check",
    "let's double check",
    "wait a minute",
    "let me start over",
    "let me try again",
    "oops"
]
\end{lstlisting}
\end{tcolorbox}

\clearpage
\subsection{Ablation Studies on \ours{} Method Design}
\subsubsection{Ablating Loss terms in SRT}
The SRT objective \lsrt{} (Section~\ref{sec:method}) combines two terms: the first term \lrev{} trains the model to produce a revision or rephrase conditioned on the outcome; the second term \lgen{} trains it to generate better responses by actively evaluating its current response and self-correcting. We ablate this by training with each term alone. If the self-correction term is redundant, removing it should not hurt Phase~2 performance; if the conditioned revision term is redundant, the model should still learn to revise from the self-correction objective alone.

As shown in \Cref{tab:phase1-ablation}, both terms are necessary, and neither alone recovers the full effect of \sft{}. Using only \lgen{} preserves relatively strong first-attempt generation, but its revised-attempt improvement is much smaller, with the correction rate dropping from 15.0\% to 7.2\%. This suggests that self-correction behavior does not emerge reliably from the generation objective alone. In contrast, using only \lrev{} yields a higher correction rate of 12.1\%, indicating that the model does learn to revise when explicitly trained to do so, but this comes at a clear cost to generation quality: both average generation accuracy and first-attempt accuracy fall substantially relative to the full \sft{} objective. Taken together, these results show that the two terms play complementary roles. \lrev{} is important for directly eliciting revision behavior, while \lgen{} helps transfer that behavior back into stronger standalone generation. Their combination is what enables \sft{} to simultaneously improve first-attempt performance and unlock effective self-revision. This complementarity is consistent with previous work 
\citep{kumar2024traininglanguagemodelsselfcorrect}, which similarly finds that self-correction does not emerge from a correction-focused signal alone, but requires preserving the model’s underlying generation ability while separately training revision behavior.

% \sk{I think the SCoRE paper also discusses the importance of this mixing (will double check), in which case we can cite SCoRE + mention their observation is consistent with ours?}

 \def\changesize#1{\fontsize{#1}{10.5pt}\selectfont}
\begin{table}[h]
\centering
\setlength{\tabcolsep}{4pt}
\small
\begin{tabular}{lcccc}
\toprule
\multirow{3}{*}{\textbf{Method}}  & \multirow{3}{*}{\textbf{Average Generation}}& \multicolumn{3}{c}{\textbf{Generate-then-Revise on AIME24}} \\
\cmidrule(lr){3-5}
&\multirow{2}{*}{\textbf{Accuracy (\%)}}&{\changesize{8.3}First Attempt}  & {\changesize{8.3}Revised Attempt} & {\changesize{8.3}Correction} \\
&  & {\changesize{8.3}Accuracy (\%)} & {\changesize{8.3}Accuracy (\%)} & {\changesize{8.3}Rate (\%)} \\

\midrule
Base Model & 49.8 & 59.6 & 60.7 & 2.7 \\
\sft{}  & 57.6 & 66.7 & 71.7 & {15.0} \\
\rowcolor{lightyellow!50}
\textbf{\sft{} (\lgen{} Only)} & 56.4 & 65.4 & 67.9 & 7.2 \\
\rowcolor{lightblue!80}
\textbf{\sft{} (\lrev{} Only)} & 52.2 & 62.1 & 66.7 & 12.1 \\
\bottomrule
\end{tabular}
\caption{\textbf{Ablation of the two loss terms in the \sft{} objective.} We train \qwen{} with \sft{} using only \lgen{} or only \lrev{}, and evaluate both overall generation quality and self-revision ability. Both terms are important: using either term alone underperforms the full \sft{} objective in average generation accuracy and in generate-then-revise performance. In particular, \lgen{} alone yields relatively weak correction after revision, while \lrev{} alone improves correction rate but leads to substantially worse first-attempt and overall generation accuracy, showing that the two terms play complementary roles in strengthening both generation and self-revision.}
\label{tab:phase1-ablation}
\end{table}

\subsubsection{Applying \opsd{} Phase Directly to the Base Model}
\Cref{tab:phase2-ablation} provides an ablation study showing what happens when we remove the \sft{} phase and apply the \opsd{} phase directly to the base model. Although this Phase-2-only variant slightly improves average generation accuracy from 49.8 to 51.4, its Generate-then-Revise (see \Cref{sec:srt-results}) behavior remains weak: first-attempt accuracy increases only modestly from 59.6 to 61.2, revised-attempt accuracy rises from 60.7 to 62.2, and the correction rate is only 2.6\%, which is nearly identical to the base model's 2.7\%. In contrast, models that include the \sft{} phase achieve much stronger revision gains, with correction rates of 15.0\% for \sft{} and 16.7\% for \ours{}. These results indicate that directly applying preference optimization to a model without prior self-revision training does not meaningfully improve its ability to revise incorrect solutions. Instead, the \sft{} phase appears to be a necessary prerequisite that first elicits self-revision behavior, after which the \opsd{} phase can effectively refine and strengthen it.
\def\changesize#1{\fontsize{#1}{10.5pt}\selectfont}
\begin{table}[h]
\centering
\setlength{\tabcolsep}{4pt}
\small
\begin{tabular}{lcccc}
\toprule
\multirow{3}{*}{\textbf{Method}}  & \multirow{3}{*}{\textbf{Average Generation}}& \multicolumn{3}{c}{\textbf{Generate-then-Revise on AIME24}} \\
\cmidrule(lr){3-5}
&\multirow{2}{*}{\textbf{Accuracy (\%)}}&{\changesize{8.3}First Attempt}  & {\changesize{8.3}Revised Attempt} & {\changesize{8.3}Correction} \\
&  & {\changesize{8.3}Accuracy (\%)} & {\changesize{8.3}Accuracy (\%)} & {\changesize{8.3}Rate (\%)} \\

\midrule
Base Model & 49.8 & 59.6 & 60.7 & 2.7 \\
\sft{}  & 57.6 & 66.7 & 71.7 & {15.0} \\
\ours{} & 60.3 & 68.3 & 73.6 & {16.7} \\
\rowcolor{lightblue!80}
\textbf{\ours{} (Phase 2 Only)} & 51.4 & 61.2 & 62.2 & 2.6\\
\bottomrule
\end{tabular}
\caption{\textbf{Ablation of applying \ours{} without the \sft{} phase.} We train a Phase-2-only variant on \qwen{} by running \ours{} directly on the base model, without first inducing self-revision through \sft{}. This variant shows only marginal improvement over the base model in both overall generation accuracy and generate-then-revise performance, and its correction rate remains nearly unchanged. In contrast, the full two-phase method substantially improves both first-attempt generation and revision effectiveness. These results suggest that the \sft{} phase is necessary to first elicit self-revision capability, which \ours{} can then refine and distill into stronger reasoning.}
\label{tab:phase2-ablation}
\end{table}

\subsubsection{Ablating Data Split Between Phases}
Our method uses a fixed overall data budget that must be divided between the \sft{} phase and the \opsd{} phase. This creates a natural trade-off: allocating more data to \sft{} may produce a stronger reviser, but leaves fewer examples for Phase 2 distillation; allocating too little data to \sft{} may instead yield a reviser that is not strong enough to provide useful supervision. To study this trade-off, we vary the split between the two phases while keeping the total amount of training data fixed, and report both the intermediate \sft{} model accuracy and the final \ours{} model accuracy in \Cref{tab:data-split-ablation}.

\def\changesize#1{\fontsize{#1}{10.5pt}\selectfont}
\begin{table}[h]
\centering
\setlength{\tabcolsep}{4pt}
\small
\begin{tabular}{cccc}
\toprule
\multicolumn{2}{c}{\textbf{Data Split}} & \multicolumn{2}{c}{\textbf{Average Performance (\%)}} \\
\cmidrule(lr){1-2} \cmidrule(lr){3-4}
\textbf{\sft{}} & \textbf{\opsd{}} & \textbf{\sft{} Model} & \textbf{\ours{} Model} \\
\midrule
6K & 9K & 57.6 & 60.3 \\
9K &  6K& 57.8 & 59.1 \\
7.5K & 7.5K & 57.8 & 59.8 \\
\bottomrule
\end{tabular}
\caption{\textbf{Ablation of data split between \sft{} and \opsd{} phases.} On \qwen{}, we vary how the training data is allocated across the two phases while keeping the total data budget fixed. Allocating more data to \sft{} slightly improves \sft{} model performance, but does not lead to the best final \ours{} accuracy. The best overall result is achieved by assigning more data to \opsd{}, suggesting that once \sft{} has sufficiently elicited self-revision capability, additional data is more effectively used in \opsd{} phase to distill and refine this behavior into stronger generation.}
\label{tab:data-split-ablation}
\end{table}
\clearpage
\section{SDFT with Thinking Models}
\label{app:opsd-thinking}
 
In Section \ref{sec:discussion} we noted that extending self-distillation to \emph{thinking
models}---where the student generates long, exploratory chains of thought
before reaching an answer---is a natural but non-trivial next step.
Below,
we provide concrete evidence for this claim by applying SDFT to
Qwen3-4B with thinking enabled during the student's on-policy
rollouts.
 
\paragraph{Setup.}
We use the \texttt{Ashkchamp/Openthoughts\_math\_filtered\_30K} dataset,
taking the \texttt{solution} column as the privileged information supplied
to the teacher.  Training runs for one epoch with learning rate
$5\!\times\!10^{-6}$ and batch size~64.  The sole experimental variable is
the \texttt{enable\_thinking} flag, which controls whether the student generates with Qwen3's native thinking mode turned on or off
\emph{during SDFT training}.
%---that is, whether the student's on-policy
%rollouts (and the corresponding teacher distributions) are produced in
%thinking mode.  
At evaluation, thinking is always enabled for all methods;
the model may generate up to 38K completion tokens, and we sample with
temperature~$0.6$ and top-$p = 0.95$, reporting avg@16.
 
\paragraph{Results.}
Table~\ref{tab:opsd-thinking} shows that enabling thinking during
training degrades the base model on every benchmark
we tested: $-9.8$ points on AIME24, $-10.8$ on AIME25, and $-8.3$ on
HMMT25.  Disabling thinking during training largely preserves (and on HMMT25 slightly
improves) the base model's accuracy.
 
\begin{table}[h]
\centering
\vspace{4pt}
\begin{tabular}{lccc}
\toprule
\textbf{Method} & \textbf{AIME24} & \textbf{AIME25} & \textbf{HMMT25} \\
\midrule
Qwen3-4B  & 0.735 & 0.647 & 0.458 \\
\midrule
\; + SDFT (\texttt{enable\_thinking=True})
  & \textcolor{darkbluegreen}{0.637} & \textcolor{darkbluegreen}{0.539} & \textcolor{darkbluegreen}{0.375} \\
\; + SDFT (\texttt{enable\_thinking=False})
  & 0.733 & 0.616 & 0.466 \\
\bottomrule
\end{tabular}
\caption{%
  Effect of the student's thinking mode on SDFT training
  (Qwen3-4B, avg@16).  \texttt{enable\_thinking} refers to
  whether thinking is active during training; all models are evaluated
  with thinking enabled.  Red entries denote degradation relative
  to the base model.}
\label{tab:opsd-thinking}
\end{table}

%\sk{can also add now where enable\_thinking=True and the thinking trace for privileged answer is also in the context; this setting makes performance plummet even more, by almost an additional 10\% compared to just enable\_thinking=True} \\
%\sk{can rerun with the r1 math dataset if there is time}
\clearpage
%\section{Early Experiments on Countdown with Qwen2.5-7B}
\section{Self-Revision Design Choices}
\label{app:countdown}

We present complementary experiments on Countdown, a constrained arithmetic reasoning task, using Qwen2.5-7B. Although the setup differs from the main experiments (different base model, task domain, and pipeline configuration), three findings align with and corroborate the design principles of \ours{}: (1) on-policy self-revision data outperforms off-policy teacher data, (2) correctness filtering of revision traces is essential, and (3) self-revision training provides a stronger foundation for subsequent training than standard SFT does.

\subsection{Effect of Training Data Source}
\label{app:countdown-sources}
 
We compare several training data sources for supervised fine-tuning on
Countdown.  For each of the 8K original SFT questions, we sample $N$ solutions per question and
optionally filter for correctness. Table~\ref{tab:countdown-sources} reports the results.
 
\begin{itemize}
\item \textbf{LLaMA-70B}: $N\!=\!16$, pass@$16 = 81\%$, filtered
      $\rightarrow$ 6.5K pairs.
\item \textbf{Qwen2.5-7B} (self-generated): $N\!=\!16$,
      pass@$16 = 78.7\%$, filtered $\rightarrow$ 6.2K pairs.
\item \textbf{GPT-4o}: $N\!=\!5$, pass@$5 = 67.5\%$, filtered
      $\rightarrow$ 5.4K pairs.
\item \textbf{SFT on GRPO-Qwen2.5-7B data} (on-policy): GRPO reaches
      pass@$1 = 91\%$ after RL; we distill its generations into
      6.5--8K SFT pairs without filtering.
\item \textbf{Self-Revision}: Qwen2.5-7B critiques and refines the previous
      response attempt; unfiltered (6.5K) and filtered
      (5.9K) variants.
\end{itemize}

\begin{table}[h]
\centering
\vspace{4pt}
\small
\begin{tabular}{lccc}
\toprule
\textbf{Training Source} & \textbf{pass@1} & \textbf{pass@2} & \textbf{pass@4} \\
\midrule
No training & 0.410 & 0.538 & 0.639 \\
\midrule
LLaMA-70B (6.5K, filtered) & 0.605 & 0.725 & 0.814 \\
Qwen2.5-7B (6.2K, filtered) & 0.552 & 0.676 & 0.768 \\
GPT-4o (5.4K, filtered) & 0.617 & 0.711 & 0.775 \\
\midrule
SFT on GRPO data (8K, unfiltered) & 0.878 & 0.904 & 0.922 \\
SFT on GRPO data (6.5K, unfiltered) & 0.883 & 0.910 & 0.919 \\
\midrule
Self-Revision (6.5K, unfiltered) & 0.529 & 0.639 & 0.717 \\
Self-Revision (5.9K, filtered) & 0.630 & 0.737 & 0.810 \\
\bottomrule
\end{tabular}
\caption{%
  Countdown pass@$k$ after fine-tuning Qwen2.5-7B on different data
  sources.  On-policy sources (SFT on GRPO data, self-revision)
  consistently outperform off-policy ones (LLaMA-70B, GPT-4o).  Among
  methods that do not require a prior RL stage, filtered self-revision
  achieves the highest pass@1.}
\label{tab:countdown-sources}
\end{table}
 
\paragraph{Findings.}
Three patterns emerge.  First, \textbf{on-policy data dominates
off-policy data}: SFT on data generated by the GRPO-trained checkpoint
(${\sim}0.88$) substantially outperforms SFT on LLaMA-70B ($0.605$) or
GPT-4o ($0.617$) data, despite comparable data sizes and correctness
filtering.  Note that GRPO itself reaches ${\sim}0.91$ via RL alone;
the ${\sim}0.88$ figure reflects SFT distillation of its outputs.
Distribution match matters more than teacher quality---the on-policy
6.5K subset performs comparably to the 8K variant.  \ours{} also
uses the student's own reviser as the teacher.
 
Second, \textbf{self-revision outperforms first-attempt sampling}:
filtered self-revision (pass@1=$0.630$) exceeds the model's own
correctness-filtered first attempts (pass@1=$0.552$).  Both are on-policy, so
the gain isolates the value of the revision step itself.
 
Third, \textbf{correctness filtering is essential}: unfiltered
self-revision ($0.529$) underperforms the base model's first attempts,
while filtering raises it to $0.630$---a $19\%$ relative gain.  Noisy
revision traces actively hurt.  \ours{}'s Phase~1 also retains only
successful revisions (Section~2.1).

\subsection{Self-Revision as Initialization for RL}

\label{app:countdown-ood}
 
We evaluate whether self-revised targets (A$'$) improve OOD
generalization, especially when combined with RL.  We fine-tune
Qwen2.5-7B on 8K Countdown questions with either LLaMA-70B--generated
answers~(A) or self-revised answers~(A$'$), optionally followed by GRPO.
Neither A nor A$'$ are filtered for correctness.  Note that A uses
off-policy teacher answers; Table~\ref{tab:countdown-sources} shows this
is a conservative baseline, since off-policy data underperforms
on-policy data even in-distribution.  All evaluations use a 4096-token
generation budget.
 
\begin{table}[h]
\centering
\vspace{4pt}
\footnotesize
\setlength{\tabcolsep}{1.5pt}
\begin{tabular}{l cc cc cc cc cc cc cc}
\toprule
& \multicolumn{2}{c}{\textbf{AIME24}}
& \multicolumn{2}{c}{\textbf{AIME25}}
& \multicolumn{2}{c}{\textbf{Countdown}}
& \multicolumn{2}{c}{\textbf{IFBench}}
& \multicolumn{2}{c}{\textbf{IFEval}}
& \multicolumn{2}{c}{\textbf{MMLU}}
& \multicolumn{2}{c}{\textbf{MMLU-Pro}} \\
\midrule
\textbf{Method}
& @1 & @128 & @1 & @128 & @1 & @128
& @1 & @128 & @1 & @128 & @1 & @128 & @1 & @128 \\
\midrule
Base
  & .07 & .27 & .07 & .43 & .41 & .90
  & .27 & .58 & .71 & .90 & .74 & .91
  & .56 & .82 \\
GRPO (Countdown)
  & .08 & .30 & .08 & .40 & .91 & .98
  & .27 & .61 & .72 & .91 & .74 & .92
  & .56 & .84 \\
\midrule
Countdown (A)
  & .05 & .27 & .05 & .43 & .39 & .91
  & .23 & .54 & .52 & .85 & .69 & .96
  & .46 & .91 \\
Countdown (A) + GRPO
  & .06 & .23 & .07 & .40 & .92 & .98
  & .24 & .55 & .53 & .84 & .71 & .97
  & .50 & .93 \\
\midrule
Countdown (A$'$)
  & .07 & .30 & .06 & .47 & .40 & .88
  & .19 & .62 & .60 & .91 & .71 & .97
  & .50 & .94 \\
Countdown (A$'$) + GRPO
  & .07 & \textbf{.40} & .07 & \textbf{.47} & .91 & .98
  & .20 & .58 & .62 & .91 & .72 & .97
  & .52 & .93 \\
\bottomrule
\end{tabular}
\caption{%
  OOD evaluation (pass@1 / pass@128) after training on Countdown with
  LLaMA-70B answers~(A) or self-revised answers~(A$'$), optionally
  followed by GRPO.  A$'$+GRPO achieves the highest OOD scores.}
\label{tab:countdown-ood}
\end{table}

%\paragraph{Findings.}
The key result is that \textbf{self-revised targets provide a better
initialization for RL on OOD benchmarks.}  Countdown~(A$'$)+GRPO yields
pass@128 of $0.40$ on AIME24 and $0.47$ on AIME25, compared to $0.23$
and $0.40$ for Countdown~(A)+GRPO, and $0.30$ and $0.40$ for standalone
GRPO.  In-distribution Countdown performance is identical across all
GRPO variants ($0.91$/$0.98$).

%\paragraph{Comparison with GRPO.}
%GRPO reaches pass@1=0.91 in-distribution, outperforming filtered self-revision (pass@1=0.63). Self-revision SFT, while the strongest non-RL method that does not require a prior RL stage, does not close this gap on Countdown. The two setups differ in base model and training recipe, so direct comparison is limited. However, self-revision provides a stronger initialization for subsequent RL than off-policy data, suggesting the two approaches are complementary rather than competing.
%In-distribution, GRPO substantially outperforms even the best self-revision SFT variant (pass@1=0.91 vs. 0.63; Tables \ref{tab:countdown-ood} and \ref{tab:countdown-sources} respectively). The two setups differ in base model and training recipe, so direct comparison is limited. However, the OOD results above suggest the two approaches are complementary rather than competing.

%This comparison is conservative in two respects.  First, A uses
%LLaMA-70B answers (off-policy), not the model's own first attempts;
%Table~\ref{tab:countdown-sources} shows that self-revision outperforms
%even on-policy first attempts ($0.630$ vs.\ $0.552$), so the advantage
%of A$'$ likely reflects the revision process itself.  Second, neither A
%nor A$'$ were filtered for correctness; given the $19\%$ relative gain
%from filtering in Table~\ref{tab:countdown-sources}, the gap would
%presumably widen further.

Notably, this comparison is conservative: neither A nor A' were filtered for correctness, and A uses off-policy 
%LLaMA-70B
answers rather than the model's own attempts. Despite this, self-revised targets still yield a meaningfully stronger RL initialization on OOD benchmarks (Table~\ref{tab:countdown-ood}). This suggests that the revision process itself — even without correctness filtering — carries useful signal, and that the gains would likely be larger with filtering (Table~\ref{tab:countdown-sources} shows a 19\% relative gain from filtering in-distribution).
Moreover, these experiments use GRPO rather than \ours{}'s distillation objective as the second stage, suggesting the value of self-revision as initialization is not specific to a particular second-stage method.
 
%The second-stage algorithm also differs: these experiments use GRPO
%rather than the SDFT distillation objective in SD-Zero's Phase~2.  The
%complementarity of self-revision with RL thus holds under a different
%second-stage method.

More broadly, the fact that unfiltered self-revision provides useful signal raises the question of whether self-revision training can extend to settings without verifiable rewards, where correctness filtering is unavailable. We leave this to future work.
 
%\paragraph{Tulu3 data.}
%We ran parallel experiments with Tulu3-generated training data, but the
%Tulu3 solutions contain extended thinking tokens and no correctness
%filtering was applied.  The A$'$ vs.\ A improvements were noisier,
%and consistent with the findings in Appendix~\ref{app:opsd-thinking} on the
%difficulty of supervising thinking-mode traces and with the importance of
%filtering documented above.

\end{document}